\documentclass{acmart}

\AtBeginDocument{%
  \providecommand\BibTeX{{%
    \normalfont B\kern-0.5em{\scshape i\kern-0.25em b}\kern-0.8em\TeX}}}

\setcopyright{acmcopyright}
\copyrightyear{2023}
\acmYear{2023}
\acmDOI{XXXXXXX.XXXXXXX}

\usepackage{makecell}
\usepackage{multirow}
\usepackage{enumitem}
\newtheorem{definition}{Definition}
\usepackage{mdframed}
\newmdtheoremenv[%
  backgroundcolor=gray!20,
  linecolor=red!60!black,
  linewidth=2pt,
  topline=false,
  rightline=false,
  skipabove=10pt,
  skipbelow=10pt,
  leftline=false
]{regbox}{Box}
\newcommand{\specialcell}[2][c]{%
    \begin{tabular}[#1]{@{}l@{}}#2\end{tabular}
}
 
\DeclareMathOperator*{\argmin}{argmin} 

\newcommand\revise[1]{\textcolor{black}{#1}}
\newcommand{\eg}{\emph{e.g.}}
\newcommand{\ie}{\emph{i.e.}}

\newcommand{\Entity}{\textbf{\textsc{Entity}}}
\newcommand{\Gene}{\textbf{\textsc{Ge}}}
\newcommand{\Protein}{\textbf{\textsc{Prt}}}
\newcommand{\Molecule}{\textbf{\textsc{Mol}}}
\newcommand{\Cell}{\textbf{\textsc{Cell}}}
\newcommand{\Drug}{\textbf{\textsc{Drug}}}
\newcommand{\Target}{\textbf{\textsc{Tgt}}}
\newcommand{\Predict}{\textbf{\textsc{Pred}}}
\newcommand{\Pair}{\textbf{\textsc{Pair}}}
\newcommand{\Action}{\textbf{\textsc{Act}}}

\begin{document}

\title{Knowledge-augmented Graph Machine Learning for Drug Discovery: A Survey}

\author{Zhiqiang Zhong}
\affiliation{%
  \institution{Aarhus University}
  \city{Aarhus}
  \country{Denmark}}
\email{zzhong@cs.au.dk}

\author{Anastasia Barkova}
\affiliation{%
  \institution{WhiteLab Genomics}
  \city{Paris}
  \country{France}}
\email{barkova.anastasia@gmail.com}

\author{Davide Mottin}
\affiliation{%
  \institution{Aarhus University}
  \city{Aarhus}
  \country{Denmark}}
\email{davide@cs.au.dk}

\renewcommand{\shortauthors}{Zhong et al.}

\begin{abstract}
The integration of \emph{Artificial Intelligence} into the field of drug discovery has been a growing area of interdisciplinary scientific research.
As of late, \emph{Graph Machine Learning} (GML) has gained considerable attention for its exceptional ability to model graph-structured biomedical data and investigate their properties and functional relationships. 
Despite extensive efforts, GML methods still suffer from several deficiencies, such as the limited ability to handle supervision sparsity and provide interpretability in learning and inference processes. 
In response, recent studies have proposed integrating external biomedical knowledge into the GML pipeline to realise more precise and interpretable drug discovery with scarce training data. 
Nevertheless, a systematic definition for this burgeoning research direction is yet to be established. 
This survey formally summarises \emph{Knowledge-augmented Graph Machine Learning} (KaGML) for drug discovery and organises collected KaGML works into four categories following a novel-defined taxonomy. 
We also present a comprehensive overview of long-standing drug discovery principles and provide the foundational concepts and cutting-edge techniques for graph-structured data and knowledge databases.
To facilitate research in this promptly emerging field, we share collected practical resources that are valuable for intelligent drug discovery and provide an in-depth discussion of the potential avenues for future advancements. 

\end{abstract}




\maketitle

\section{Introduction} 
\label{sec:introduction}
\begin{figure}[!ht]
\centering
\includegraphics[width=.7\linewidth]{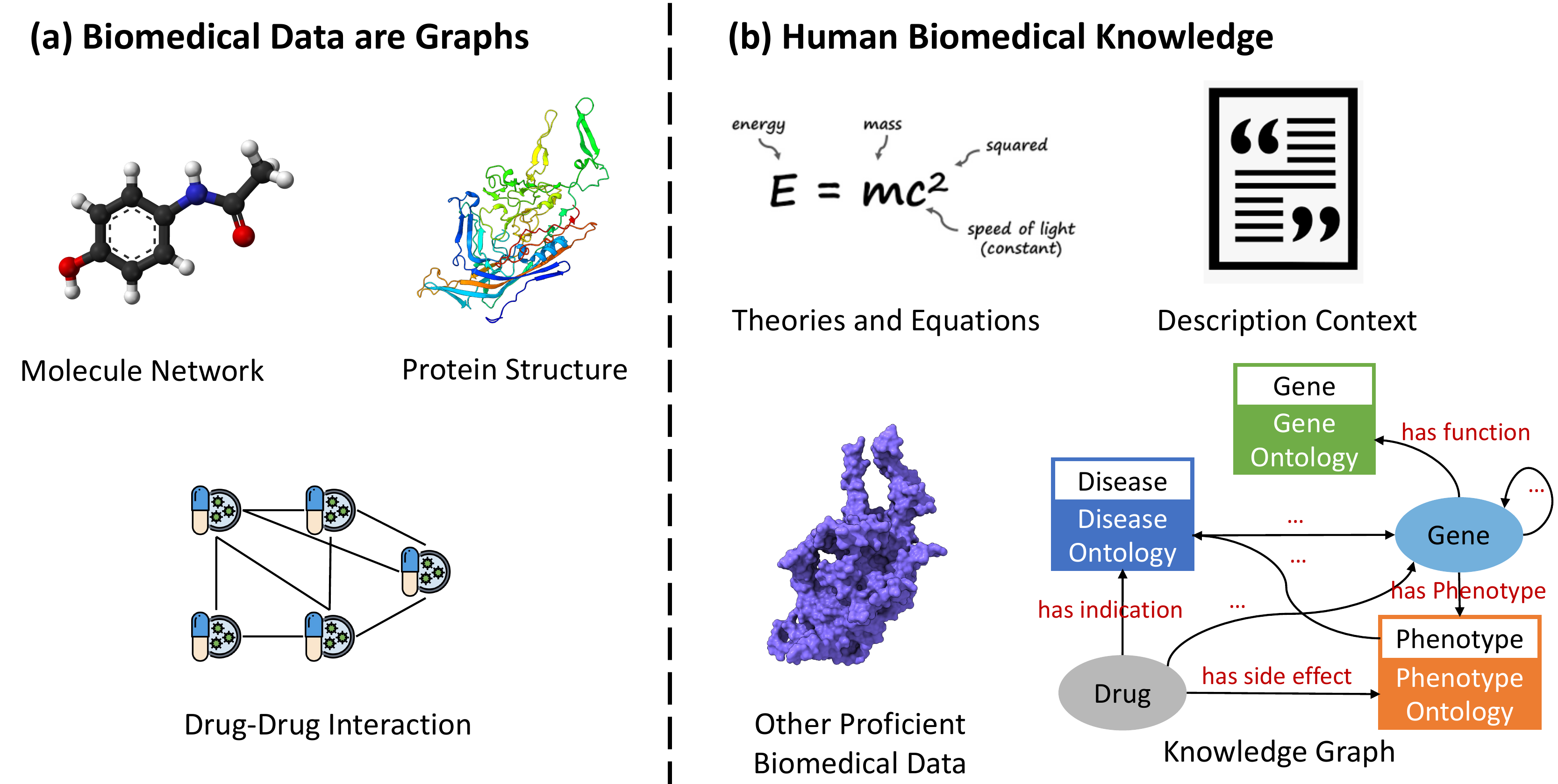}
\caption{
\revise{
Illustration of real-world biomedical data in the form of graphs (a) and examples of human biomedical knowledge (b).
}
}
\Description{}
\label{fig:illustration_of_real_world_biomedical_data}
\vspace{-5mm}
\end{figure}

Drug discovery and development have been one of the most prominent and challenging research tasks for decades~\cite{V90,K00,HG07}. 
Prior to a drug being marketed and distributed to patients, it must undergo a multitude of research validations.
From initial early drug discovery to preclinical development, and subsequent to clinical trials and final regulatory approval, it usually takes $10$-$15$ years and costs around $2$ billion US dollars~\cite{DGH16,MG17,WML20}. 
%
%
To reduce the financial burden and increase the success rate, researchers have been working on accelerating drug discovery by taking advantage of \textit{Artificial Intelligence} (AI) techniques~\cite{GHS16,ZTHZ17,DHJKOS18,VAM21,HFGZRLCXSZ22}.
Technological advances now allow for the creation of vast amounts of data in areas such as genomics, proteomics, and imaging, which can be used to inform the drug discovery process~\cite{DHJKOS18,CZA18,SCNNSGCAMJ22}. 

Biomedical data is highly interconnected~\cite{GDJSRLHVRTRBBT21,LHZ22} and can be easily represented as \textit{graphs} (or \textit{networks}), which have a variety of applications at different stages of the drug discovery and development process. 
For instance, as illustrated in Figure~\ref{fig:illustration_of_real_world_biomedical_data}-(a), biomedical data can be hierarchically represented as graphs. 
\revise{
Starting from the molecular level, atoms can be represented as nodes, and chemical bonds/Euclidean distance between atoms as edges of (2D or 3D) molecular graphs~\cite{TLM94,E00}.
}
On the macro-molecule level, interactions (edges) between amino acid residues (nodes) organise as (2D or 3D) protein graphs~\cite{Z08,MCSHPZS11}. 
At the compound level, edges in the drug-drug interaction (DDI) network can indicate chemical interactions (edges) between drugs (nodes) measured by long-term clinical screens~\cite{PS90,ZZZH09}. 

\textit{Graph Machine Learning} (GML), a class of AI methods, has been proposed to investigate graph-structured data.
\revise{
The essential idea of GML is to learn effective feature representations of nodes (\eg, drugs in DDI networks)~\cite{KCJUBD19}, edges (\eg, relations or interactions between drug-drug or drug-disease)~\cite{GAPSHLF22}, or the (sub)graphs (\eg, molecular graphs)~\cite{DFSL22}.
}
These corresponding node-, edge- and (sub)graph-level downstream tasks can be realised based on these learned representations.
According to different representation learning mechanisms, GML approaches can be broadly categorised into \revise{``shallow'' (Section~\ref{subsubsec:shallow_grl}) and ``deep'' (Section~\ref{subsubsec:deep_grl})} classes. 
In particular, a type of deep GML method called \textit{Graph Neural Networks} (GNNs)~\cite{CWPZ19, ZCZ20, XSYAWPL21, WPCLZY21, J22}, which are deep neural network architectures specifically designed for graph-structure data, are attracting growing interest. 
GNNs iteratively update the features of graph nodes by propagating the information from their neighbouring nodes. 
These methods have already been successfully applied to a range of tasks and domains, including drug discovery~\cite{GDJSRLHVRTRBBT21,XXCJZ21,AGS21}. 

However, despite the current pace of GML in drug discovery, they suffer from several serious deficiencies, 
\revise{
including \textit{high data dependency}, whereby strong performance relies on high-qualified and expensive to obtain training data of the target application~\cite{M18,PSYTTRSCI19}, 
\textit{restricted biomedical expertise}, whereby constructed datasets are deficient in advanced domain knowledge, 
and \textit{poor generalisation}, whereby the model perform subpar on instances that have never been observed in training data~\cite{BC21}.
}
These deficiencies originate primarily from the models' data-driven nature and inability to exploit the domain knowledge effectively.
In addition, there has been an increased demand for methods that help people understand and interpret the underlying models and provide more trustworthiness. 
In an effort to mitigate the lack of interpretability and trustworthiness of certain machine learning models and to augment human reasoning and decision-making, attention has been drawn to \textit{eXplainable Artificial Intelligence} (XAI)~\cite{JGS20} and \textit{Trustworthy Artificial Intelligence} (TAI)~\cite{TLS21} approaches, that provide human-comprehensible explanations for the model's inherent mechanism and outputs. 

To address these limitations, researchers recently paid attention to a new AI paradigm, which we refer to as \textit{Knowledge-augmented Graph Machine Learning} (KaGML in short), for superior drug discovery. 
Its core idea is to integrate external human biomedical knowledge into different components of the GML pipeline to achieve more accurate drug discovery, along with user-friendly interpretations, which guarantees the expert’s knowledge is not to be substituted. 
Biomedical knowledge may exist in various forms, as shown in Figure~\ref{fig:illustration_of_real_world_biomedical_data}-(b), including formal scientific knowledge (\eg, well-established laws or theories in a domain that govern the properties or behaviours of target variables), informal experimental knowledge (\eg, well-known facts or rules extracted from longtime observations and can also be inferred through humans’ reasoning). 
The contributions of this survey are the following:
\begin{itemize}[leftmargin=*]\itemsep0em 
    \item We are the first to propose the concept of KaGML and comprehensively summarise existing work.
    The discussion between KaGML and existing other paradigms emphasises the novelty of KaGML and its promising potential for practical medical applications. 
    \item We propose a novel taxonomy of KaGML approaches according to different schemes to incorporate knowledge into the GML pipeline.
    It is easier for the readers to identify the core design of different models and locate the interesting categories (Section~\ref{sec:kagml_for_dd}).
    We created a public folder to share collected resources\footnote{\url{https://github.com/zhiqiangzhongddu/Awesome-Knowledge-augmented-GML-for-Drug-Discovery}} and will contribute to it continuously.
    \item We carefully discuss practical tools and knowledge databases that have been (or are highly possible to be) used by KaGML methods to solve practical drug discovery problems (Section~\ref{sec:practical_resources}).
    We provide a schematic representation of possible schemes to organise different knowledge databases about small molecule drugs into one KG. 
    \item We cover the methodologies not only for solving scientific problems under a computer science scenario but, more importantly, for real-world biomedical applications. 
    Our survey is hence of interest not only to AI researchers but also to biologists in different fields. 
    Section~\ref{sec:discussion_and_open_challenges} discusses promising future work for researchers from both disciplines to exploit. 
\end{itemize}

\begin{table}[!ht]
\caption{
    Summary of the keywords used in the literature search.
}
\label{table:keywords_literature_search}
\centering
\small
\begin{tabular}{l|l}
\toprule 
\textbf{Area} & \textbf{Keywords} \\
\midrule
Drug Discovery & \specialcell{Drug Discovery, Drug Design, Drug Development, Medicine Discovery, \\ Medicine Design, Medicine Development} \\
& \\
Knowledge Graph & \specialcell{Knowledge-augmented, Knowledge-aware, Knowledge-informed, Knowledge-guided, \\
Knowledge-enhanced, Knowledge-driven} \\
& \\
Graph Machine Learning & \specialcell{Graph Machine Learning, Graph Neural Network, Geometric Machine Learning} \\
\bottomrule
\end{tabular}
\end{table}

\noindent
\textbf{Search methodology.}
All studies were retrieved in one of three following ways: 
\textit{(i)} a comprehensive top-down approach that conducted an extensive search of KaGML papers from major academic databases such as Google Scholar, IEEExplore, ACM Digital Library, DBLP Computer Science Bibliography, and ScienceDirect, using keywords listed in Table~\ref{table:keywords_literature_search};
\textit{(ii)} a bottom-up approach that surveyed recent research outputs in AI conferences and workshops; 
\textit{(iii)} a thorough examination of the related work, discussion sections, and cited references from the papers obtained in steps \textit{(i)} and \textit{(ii)} to identify overlooked works.
We keyword-searched for works containing a conjunction of any of the terms summarised in Table~\ref{table:keywords_literature_search}, leading to a selection of around 1,600 articles. 
Approximately 120 were thoroughly scanned according to the criteria, and about 20 were identified directly from the related work sections.
Whenever possible, we prioritised peer-reviewed publications and major journals/conferences (\eg, Nature, Nat. Mach. Intell, Nat. Commun, NeurIPS ICML, ICLR, AAAI, KDD) to white papers or unreviewed submissions.
Studies were selected only if presenting a subsymbolic system, including some forms of incorporating biomedical knowledge into GML models for precision drug discovery and producing any explanations using background knowledge with GML models. 
The finally identified papers are summarised into different categories in Table~\ref{table:summary_paper}. 



\section{Background}
\label{sec:background}
\subsection{\revise{Drug Discovery Procedures}} 
\label{subsec:dd_procedure}
\revise{
The drug discovery process is a complex and multifaceted journey that transforms basic scientific research into effective and safe medications~\cite{HRKP11,DGH16,WML20}. 
Such a process typically begins with \emph{target identification} in which researchers pinpoint specific biological molecules or pathways involved in a disease. 
Following this, \emph{high-throughput screening}, that utilises automation, miniaturised assays, and large compound libraries to identify active compounds, antibodies, or genes that modulate a particular biomolecular pathway, rapidly test thousands to millions of compounds for potential activity against the target. 
This stage is crucial for identifying promising lead compounds. 
These compounds then undergo rigorous optimisation to enhance their efficacy, selectivity, and pharmacokinetic properties. 
Subsequent \emph{preclinical testing} involves in vitro and in vivo experiments to assess the safety and biological activity of the optimised compounds. 
Successful candidates then progress to \emph{clinical trials}, which are conducted in multiple phases to ensure their safety, efficacy, and dosage in humans. 
Technological advancements made possible high-throughput screening techniques to identify lead compounds that have the potential to give rise to a drug candidate. 
However, those conventional approaches lack precision and are demanding in terms of human and financial resources. 
Throughout the process, advanced AI technologies increasingly play a role, augmenting traditional methods and expediting the discovery of new drugs~\cite{GDJSRLHVRTRBBT21,XXCJZ21,AGS21}. 
}

\subsection{Graph Machine Learning} 
\label{subsec:gml}

\begin{figure}[!ht]
\centering
\includegraphics[width=.7\linewidth]{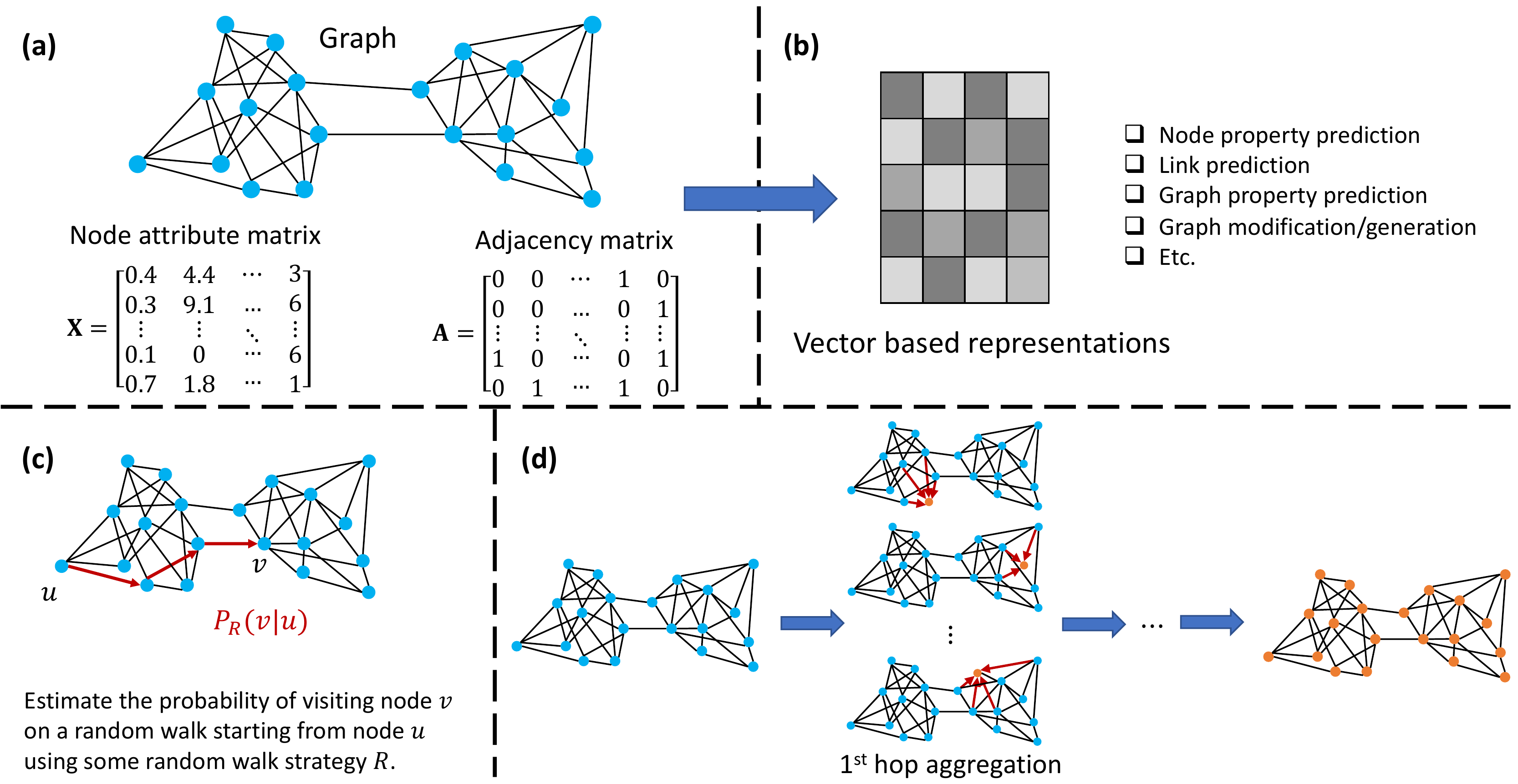}
\caption{
Toy examples of the graph and typical graph representation learning approaches.
(a): A graph can be basically represented using a node attribute matrix $\mathbf{X}$ and an adjacency matrix $\mathbf{A}$.
(b) Graph representation learning can convert a graph into a set of vectors, which record information about the graph.
(c) A toy example of random walk-based shallow GRL approaches.
(d) A toy example of GNN mechanism.
}
\Description{}
\label{fig:example_graph_and_grl}
\end{figure}

\begin{regbox} \emph{\textbf{Fundamentals of Graph Machine Learning}}
\label{box:fundamentals_graph_machine_learning}

\begin{definition}[Graph]
\label{def:graph}
A graph with $n$ nodes can be formally represented as $\mathcal{G} = (\mathcal{V}, \mathcal{E})$, which consists of $n$ nodes $v \in \mathcal{V}$ and $\vert \mathcal{V} \vert = n$.
$\mathcal{E} \subseteq \mathcal{V} \times \mathcal{V}$ denotes the set of edges, where $e_{u,v}$ denotes the edge between $u$ and $v$. 
Node attribute vector $\mathbf{x}_u \in \mathbb{R}^d$ describes side information and metadata of node $u$. 
The node attribute matrix $\mathbf{X} \in \mathbb{R}^{n \times d}$ contains attribute vectors for all nodes in the graph. 
Similarly, edge attributes $\mathbf{x}_{u,v} \in \mathbb{R}^\tau$ for edge $e_{u,v}$ can be taken together to form an edge attribute matrix $\mathbf{X}^e \in \mathbb{R}^{m \times \tau}$. 
A path from node $u_1$ to node $u_k$ is a sequence of edges $u_1 \xrightarrow{e_{1,2}} u_2 \cdots u_{k-1} \xrightarrow{e_{k-1,k}} u_k$. 
For subsequent discussion, we summarise $\mathcal{V}$ and $\mathcal{E}$ into an adjacency matrix $\mathbf{A} \in \{0, 1\}^{n \times n}$, where each entry $\mathbf{A}_{u,v}$ is 1 if $e_{u,v}$ exists, and 0 otherwise.
An example graph and its node attribute matrix and adjacency matrix are shown in Figure~\ref{fig:example_graph_and_grl}-(a).
\end{definition}


\begin{definition}[$\lambda$-hop Neighbourhood and Subgraph]
\label{def:hop_neighbourd}
The $\lambda$-hop neighbourhood of node $u$ is the set of nodes that are at a distance less than or equal to $\lambda$ from node $u$, that is, $\mathcal{N}^\lambda(u) = \{v \; \vert \; 0 < d(u, v) \leq \lambda\}$ where $d (\cdot)$ denotes the shortest path distance.
Subgraph $\mathcal{S}^{\lambda}(u) = (\mathcal{V}^{\lambda}_\mathcal{S}, \mathcal{E}^{\lambda}_\mathcal{S})$ is a subset of a graph $\mathcal{G}$, where $\mathcal{V}^{\lambda}_\mathcal{S}:= (\mathcal{N}^{\lambda}(u) \cup \{u\} ) $ and $\mathcal{E}^{\lambda}_\mathcal{S}:= ((\mathcal{V}^{\lambda}_\mathcal{S} \times \mathcal{V}^{\lambda}_\mathcal{S}) \cap \mathcal{E})$. 
\end{definition}


\noindent
\textbf{Graph analysis in Artificial Intelligence era.}
To process the graph-structured data, \emph{Graph Machine Learning} (GML)~\cite{XSYAWPL21} is designed as a predominant approach to finding effective data representations from graph data. 
The principal target of GML is to extract the desired features of a graph as informative representations that can be easily used by downstream tasks such as node-level, edge-level and graph-level, analysis, classification and regression tasks. 
Traditional GML approaches mainly rely on handcrafted features, including graph statistics~\cite{BKT13}, (\eg, degree, centrality and clustering coefficient), kernel functions~\cite{VSKB10} and experts designed features~\cite{LK07}.
However, traditional GML models are built on top of manually designed or processed feature sets.
The developed feature extractors are often not transferable and need to be designed specifically for each dataset and task.
These conventional approaches often suffer from practical limits on large-scale graphs with rich node and edge attributes.
Recently, \emph{graph representation learning}~\cite{CWPZ19,ZCZ20,XSYAWPL21} emerged to be a promising direction. 

\begin{definition}[Graph Representation Learning]
\label{def:graph_representation_learning}
Given a graph $\mathcal{G} = (\mathcal{V}, \mathcal{E})$, the task of graph representation learning (or equivalently graph embedding) is to learn a mapping function to generate vector representations for graph elements $f_{GRL}: \mathcal{G} \to  \mathbf{Z}$, such that the learned representations ($\mathbf{Z}$), \ie, \emph{embeddings}, can capture the structure and semantics of graph.
The mapping function's effectiveness is evaluated by applying $\mathbf{Z}$ to different downstream tasks.
A toy example shows the pipeline in Figure~\ref{fig:example_graph_and_grl}-(a)-(b).
\end{definition}

Depending on the \emph{Graph Representation Learning} (GRL) model's inherent architecture, existing GRL methods can be categorised into ``shallow'' or ``deep'' groups. 
Shallow GRL methods comprise an embedding lookup table that directly encodes each node as a vector and is optimised during training. 
The deep GRL methods - \emph{Graph Neural Networks} (GNNs) - have recently shown promising results in modelling structural and relational data~\cite{WPCLZY21}.

\begin{definition}[Graph Machine Learning Training]
\label{def:gml_training}
Given a graph $\mathcal{G}=(\mathcal{V}, \mathcal{E})$ and a graph representation learning model $f_{GRL}$.
The graph machine learning training mechanism can be defined as finding the parameters $\theta$ that minimise the differences between predictions and training signals:
\begin{equation}
    \argmin_{\theta} \; \mathcal{L}(f_{GRL_{\theta}}(\mathcal{G}), \mathbf{Y}) 
\end{equation}
where $\theta$ represents the trainable parameters of $f_{GRL}$, $\mathcal{L}$ is the loss function to measure the differences between predictions and training signals. 
The training signal $\mathbf{Y}$ can be a discrete one-hot/multi-hot vector (classification) or a continuous vector (regression, link prediction). 
Different loss functions (\eg, cross-entropy loss, mean squared error loss, discriminator loss, etc.) and optimisation approaches (Adam optimiser, SGD optimiser, Adagrad optimiser, etc.) can be adopted according to the requirements of $f_{GRL}$ and the downstream tasks~\cite{LBH15,GBC16}. 
\end{definition}


\end{regbox}

\subsubsection{Shallow Graph Representation Learning} 
\label{subsubsec:shallow_grl}

Shallow GRL methods comprise an embedding lookup table $\mathbf{Z}$ which directly encodes each node $v$ as a vector $\mathbf{z}_v$ and is optimised during training. 
Within this group, several Skip-Gram~\cite{MSCCD13}-based NE methods have been proposed, such as DeepWalk~\cite{PAS14} and node2vec~\cite{GL16}, as well as their matrix factorisation interpretation NetMF~\cite{QDMLWT18}, LINE~\cite{TQWZYM15} and PTE~\cite{TQM15}. 
As depicted in Figure~\ref{fig:example_graph_and_grl}-(c),
DeepWalk generates walk sequences for each node on a network using truncated random walks and learns node representations by maximising the similarity of representations for nodes that occur in the same walks, thus preserving neighbourhood structures. 
Node2vec increases the expressivity of DeepWalk by defining a flexible notion of a node's network neighbourhood and designing a second-order random walk strategy to sample the neighbourhood nodes;
LINE is a special case of DeepWalk when the size of the node's context is set to one;
PTE can be viewed as the joint factorisation of multiple networks' Laplacians~\cite{QDMLWT18}. 
To capture the structural identity of nodes independent of network position and neighbourhood's labels, struc2vec~\cite{RSF17} encodes structural node similarities at different scales.
Despite their relative success, shallow GRL methods often ignore the richness of node attributes and only focus on the network structural information, which hugely limits their performance.

\subsubsection{Deep Graph Representation Learning with Neural Network} 
\label{subsubsec:deep_grl}

\textit{Graph Neural Networks} (GNNs) are a class of neural network models suitable for processing graph-structured data. 
They use the graph structure $\mathbf{A}$ and node features $\mathbf{X}$ to learn a representation vector of a node $\mathbf{z}_v$, or the entire graph $\mathbf{z}_{\mathcal{G}}$. 
Modern GNNs~\cite{ZLP23} follow a common idea of a recursive neighbourhood aggregation or message-passing scheme, where we iteratively update the representation of a node by aggregating representations of its neighbouring nodes. 
After $\ell$ iterations of aggregation or message-passing, a node's representation captures the graph structural information within $\ell$-hop neighbourhood.
Thus, we can formally define $\ell$-th layer of a GNN as:
\begin{equation}
\label{eq:gnn_aggregate_combine}
\begin{aligned}
\mathbf{m}^{(\ell)}_a &= \textsc{Aggregate}^{N}(\{\mathbf{A}_{u,v}, \, \mathbf{h}^{(\ell-1)}_u \, \vert \, u \in \mathcal{N}(v) \}), \\
\mathbf{m}^{(\ell)}_v &= \textsc{Aggregate}^{I}(\{\mathbf{A}_{u,v} \, \vert \, u \in \mathcal{N}(v) \}) \, \mathbf{h}^{(\ell-1)}_v, \\
\mathbf{h}^{(\ell)}_v &= \textsc{Combine}(\mathbf{m}^{(\ell)}_a, \mathbf{m}^{(\ell)}_v)
\end{aligned}
\end{equation}
where $\textsc{Aggregate}^{N}(\cdot)$ and $\textsc{Aggregate}^{I}(\cdot)$ are two parameterised functions to learn during training process. 
$\mathbf{m}^{(\ell)}_a$ is aggregated message from node $v$'s neighbourhood nodes $\mathcal{N}(v)$ with their structural coefficients, 
and $\mathbf{m}^{(\ell)}_v$ is the residual message from node $v$ after performing an adjustment operation to account for structural effects from its neighbourhood nodes.
After, $\mathbf{h}^{(\ell)}_v$ is the learned as representation vector of node $v$ with combining $\mathbf{m}^{(\ell)}_a$ and $\mathbf{m}^{(\ell)}_v$, 
with a $\textsc{Combine}(\cdot)$ function, at the $\ell$-th iteration/layer.
Note that we initialise $\mathbf{h}^{(0)}_v = \mathbf{x}_v$ and the final learned representation vector after $L$ iterations/layers $\mathbf{z}_v = \mathbf{h}^{(L)}_v$.
We illustrate the learning mechanism of GNN models in Figure~\ref{fig:example_graph_and_grl}-(d). 
In addition, in terms of the representation of an entire graph ($\mathbf{z}_{\mathcal{G}}$), we can apply a $\mathrm{READOUT}$ function to aggregate node representations of all nodes of the graph $\mathcal{G}$, as 
\begin{equation}
\label{eq:gnn_readout}
    \mathbf{z}_{\mathcal{G}} = \mathrm{READOUT}(\{\mathbf{z}_{v} \; \vert \; \forall \mathbf{z}_{v} \in \mathcal{V} \})
\end{equation}
where $\mathrm{READOUT}$ can be a simple permutation invariant function such as summation or a more sophisticated graph-level pooling function~\cite{YYMRHL18,MWAT19,LLK19,RST20,YJ20}.

\begin{table}[!ht]
\caption{
Define different GNN variants according to Equation~\ref{eq:gnn_aggregate_combine}.
}
\label{table:summary_gnn_variants}
\centering
\footnotesize
\begin{tabular}{ |l|c|c|c| }
\hline
\textbf{GNN Model} & \textbf{$\textsc{Aggregate}^{N}(\cdot)$} & \textbf{$\textsc{Aggregate}^{I}(\cdot)$} & \textbf{$\textsc{Combine}(\cdot)$} \\
\hline
GCN~\cite{KW17} & $\sum\limits_{u \in \mathcal{N}(v)}\frac{\mathbf{W}^{(\ell)} \mathbf{h}^{(\ell-1)}_u}{\sqrt{\vert \mathcal{N}(u) \vert \vert \mathcal{N}(v) \vert }}$ & $\frac{\mathbf{W}^{(\ell)} \mathbf{h}^{\ell-1}_v}{\sqrt{\vert \mathcal{N}(v)\vert \vert \mathcal{N}(v) \vert }}$ & $\sigma(\textsc{Sum}(\mathbf{m}^{(\ell)}_a, \mathbf{m}^{(\ell)}_v))$ \\
\hline
GraphSAGE~\cite{HYL17} & $\textsc{Agg}(\{\mathbf{h}^{(\ell-1)}_u \, \vert \, u \in \mathcal{N}(v) \})$ & $\mathbf{h}^{(\ell-1)}_v$ & $\sigma(\mathbf{W}^{(\ell)} \cdot \textsc{Concat}(\mathbf{m}^{(\ell)}_a, \mathbf{m}^{(\ell)}_v))$ \\
\hline
GAT~\cite{VCCRLB18} & $\sum\limits_{u \in \mathcal{N}(v)}\alpha_{u,v}\mathbf{W}^{(\ell)}\mathbf{h}^{(\ell-1)}_u$ & $\alpha_{vv}\mathbf{W}^{(\ell)}\mathbf{h}^{(\ell-1)}_v$ & $\sigma(\textsc{Sum}(\mathbf{m}^{(\ell)}_a, \mathbf{m}^{(\ell)}_v))$ \\
\hline
GIN~\cite{XHLJ19} & $\sum\limits_{u \in \mathcal{N}(v)}\mathbf{h}^{(\ell-1)}_u$ & $(1+\epsilon)\mathbf{h}^{(\ell-1)}_v$ & $\textsc{MLP}_{\theta}(\textsc{Sum}(\mathbf{m}^{(\ell)}_a, \mathbf{m}^{(\ell)}_v)))$ \\
\hline
\end{tabular}
\end{table}

\revise{
Following the general structure of GNNs as defined in Equation~\ref{eq:gnn_aggregate_combine}, we can further generalise the existing GNNs as variants.
For instance, several classic and popular GNNs can be summarised as Table~\ref{table:summary_gnn_variants}.
We only survey prior and concurrent work on GML related to our contributions where necessary. 
For an overview of recent variants and applications of GML, we recommend the comprehensive survey articles~\cite{CZC18,CWPZ19,ZCZ20,WPCLZY21,J22}. 
Note that recent work explicitly discusses the close affiliation between one popular framework, \ie, Transformer, and GNNs~\cite{KNMCLLH22}. 
Hence, we treat Transformers as a special form of GNNs in this paper. 
}

\subsubsection{Applications of Graph Machine Learning}
\label{subsubsec:application_graph_machine_learning}

Typical example graph analysis applications include node role identification~\cite{BCM11}, personalised recommendation~\cite{ZZP20}, social healthcare~\cite{BM21}, academic network analysis~\cite{TZYLZS08}, graph classification~\cite{ZCNC18} and epidemic trend study~\cite{CZP21}. 
Next, we formally represent several example analysis tasks that are of great importance to the topics covered in this survey.

\begin{itemize}[leftmargin=*]\itemsep0em 
    \item \textbf{Node property prediction:} In real-world graphs ($\mathcal{G}$), nodes ($\mathcal{V}$) are often associated with semantic labels ($\mathcal{Y}$) relevant to certain information about them, such as gender, age, affiliation etc.
    However, these graphs are often partially (or sparsely) labelled or even unlabelled due to the high cost of node selection and labelling. 
    With partially labelled nodes (supervised settings) ($\mathcal{Y}_{Train}$), the node property prediction task aims to identify labels for the rest of unlabelled nodes ($\mathcal{Y}_{Test}$) by leveraging connectivity patterns of labelled ones extracted from the graph structure.
    \item \textbf{Link prediction:} The topology encoded in the adjacency matrix ($\mathbf{A}$) of a graph ($\mathcal{G}$) is not always complete as partial links between nodes can be temporally missing. 
    Link prediction aims to infer the presence of emerging links ($\mathbf{A}^{+}$) in the future based on node attributes and observed graph structure and evolution mechanism. 
    Since real-world graphs evolve fast and continuously, link prediction is crucial to know what could happen soon and can help recover a whole social graph when we only have an incomplete view.
    \item \textbf{Graph property prediction:} In real-world graphs ($\mathcal{G}$), nodes ($\mathcal{V}$) tend to interact with nodes having similar interests or backgrounds, thus forming different graph sets or sub-graph sets ($\mathcal{D} = \{\mathcal{G}_{1}, \mathcal{G}_{2}, \dots\}$).
    We can categorise them by finding commonalities among graphs of the same group and differences between different graph groups. 
    \item \textbf{Graph generation:} The objective is to modify a graph $\mathcal{G}$ or to generate a new graph $\mathcal{G}'$ representing a biomedical entity that is likely to have a property of interest, such as high drug-likeness. 
    The model takes a set of graphs $\mathcal{D} = \{\mathcal{G}_1, \mathcal{G}_2, \dots \}$ with such a property and learns a non-linear mapping function characterising the distribution of graphs in $\mathcal{D}$. 
    The learned distribution generates a new graph $\mathcal{G}'$ with the same property as input graphs.
\end{itemize}

\subsection{\revise{Knowledge Representation}} 
\label{subsec:knowledge_representation}
The vast amount of knowledge generated by drug discovery experiences in the past is invaluable for facilitating and accelerating new drug discovery. 
\revise{
Most human knowledge has been stored in various informal formats, \eg, different knowledge bases as shown in Figure~\ref{fig:illustration_of_real_world_biomedical_data}-(b). 
For instance, formal scientific knowledge (\eg, well-established laws or theories in a domain that govern the properties or behaviours of target variables) and informal experimental knowledge (\eg, well-known facts or rules extracted from longtime observations and can also be inferred through humans’ reasoning). 
}
Furthermore, the current advances in bioinformatics facilitated the construction of large-scale data networks (\eg, knowledge graphs) connecting biomedical knowledge such as chemical compounds, proteins, diseases and patients.
This section introduces knowledge databases and knowledge graphs. 
Finally, we present the popular techniques to export useful information from large knowledge databases. 

\begin{figure}[!ht]
\centering
\includegraphics[width=.75\linewidth]{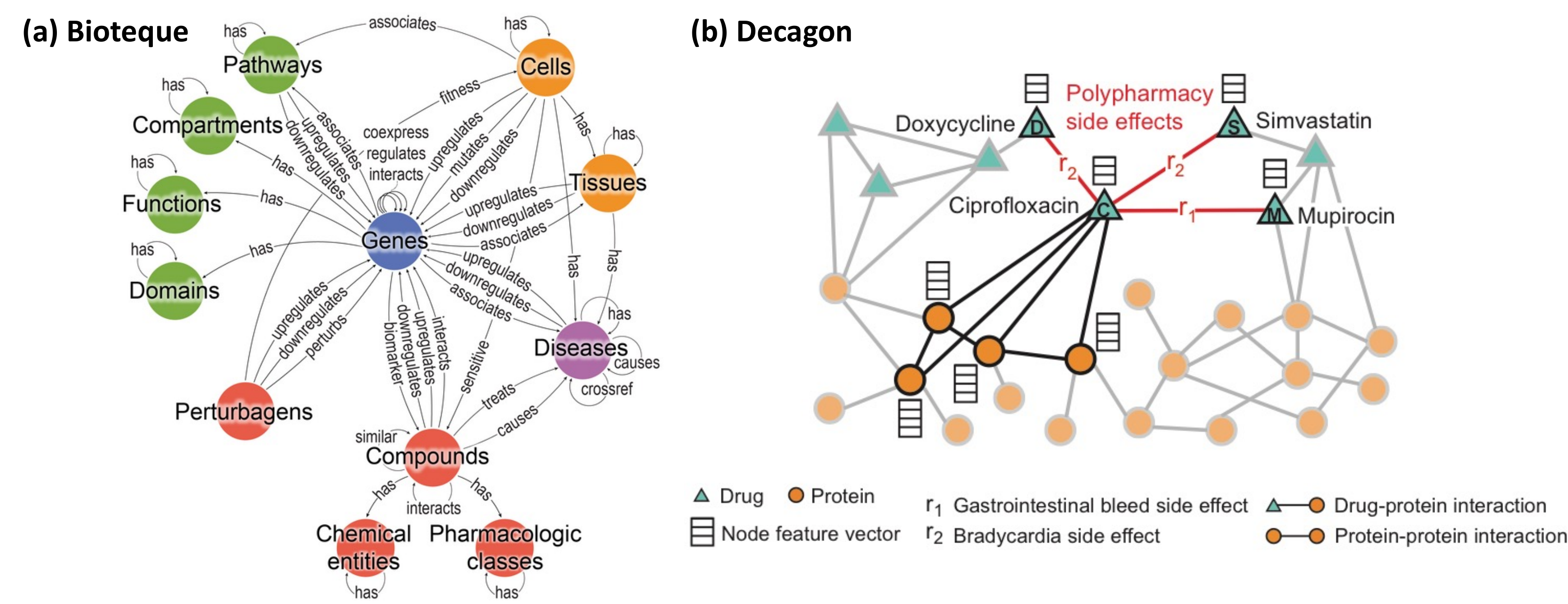}
\caption{
Illustration of example knowledge databases.
(a) Metagraph of the Bioteque~\cite{TFBLA22}, showing all the entities and the most representative associations between them.
(b) An example graph of polypharmacy side effects derived from genomic and patient population data~\cite{ZAL18}. 
More explanations about the biomedical terms will be presented in Section~\ref{sec:dd}. 
}
\Description{}
\label{fig:example_kg}
\end{figure}

\begin{regbox} \revise{\textit{\textbf{Fundamentals of Knowledge Representation}}}
\label{box:fundamentals_knowledge_database}

``Knowledge is the awareness of facts or as practical skills, the understanding of the world, and may also refer to familiarity with objects or situations''\footnote{\url{https://dbpedia.org/page/Knowledge}}.
Based on knowledge's representation, \citet{CGTJ22} group knowledge that has been explored in deep learning into two categories: scientific knowledge and experiential knowledge. 
Scientific knowledge represents the well-established laws or theories in a domain that govern the properties or behaviours of target variables. 
In contrast, experiential knowledge refers to well-known facts or rules extracted from long-term observations. It can also be inferred through human reasoning and is usually represented as knowledge databases. 

\begin{definition}[Knowledge Database]
\label{def:knowledge_base}
A knowledge database $\mathcal{D}$ represents the identified knowledge in a well-organised and structured format.
The appropriate database format depends on the type of knowledge.
Scientific knowledge is usually rigorously expressed using theories or mathematical equations. 
Experimental knowledge is less formal than scientific knowledge, often represented through probabilistic dependencies, logic rules or knowledge graphs~\cite{CGTJ22}. 
\end{definition}

\begin{definition}[Knowledge Graph]
\label{def:knowledge_graph}
A knowledge graph (KG) is a database represented as a directed heterogeneous graph $\mathcal{KG} = (\mathcal{V}, \mathcal{E}, \phi, \psi)$ with an entity type mapping function $\phi: \mathcal{V} \to \mathbf{Z}^{\mathcal{A}}$ and an edge type mapping function $\psi: \mathcal{E} \to \mathbf{Z}^{\mathcal{R}}$. 
Each entity $v \in \mathcal{V}$ belong to one or multiple entity types $\phi(v) \subseteq \mathcal{A}$ and each edge $e_{u,v} \in \mathcal{E}$ belongs to one or multiple relation types $\psi(e_{u,v}) \subseteq \mathcal{R}$. 
In addition, edges of a knowledge graph $\mathcal{KG}$ are subject-property-object triple facts, each edge of the form (\emph{head entity, relation, tail entity}) (denoted as $\langle h_u, r, t_v \rangle$) indicates a relationship of $r$ from entity $h_u$ to entity $t_v$. 
It can be regarded as an instance of a graph (Definition~\ref{def:graph}) with complex semantics. 
Figure~\ref{fig:example_kg} illustrates example knowledge graphs. 
\end{definition}

\begin{definition}[Knowledge Graph Representation Learning]
\label{def:knowledge_graph_representation_learning}
Similar to graph representation learning (Definition~\ref{def:graph_representation_learning}), the objective of knowledge graph representation learning (or termed as knowledge graph embedding in some papers) is to learn a mapping function to embed a knowledge graph $\mathcal{KG} = (\mathcal{V}, \mathcal{E}, \phi, \psi)$ into a low dimensional space $f_{KGRL}: \mathcal{KG} \to \mathbf{Z}$.
After the representation learning procedure, each graph component, including the entity and the relation, is represented with a $d$-dimensional vector.
The low dimensional embedding still preserves the inherent property of the graph, which can be quantified by semantic meaning or high-order proximity in the graph.
\end{definition}

\end{regbox}

\subsubsection{Applications of Knowledge Graph}
\label{subsubsec:application_of_knowledge_graph}

Many interesting applications for KGs exist, in addition to the graph analysis applications that can be performed on general graphs, such as entity classification and graph completion (link prediction). 
Some of the applications are specific to KGs,~\cite{WMWG17,JPCMP21}.
such as triple classification~\cite{TZJCA19}, relation extraction~\cite{BNSMSHK21} and question answering~\cite{ZDKSS18}. 
Besides, KG, as a practical format to store and present human knowledge, has been widely adopted as an information resource to assist various tasks~\cite{ZHLJSL19,WHCLC19,LCCR19}. 

\begin{itemize}[leftmargin=*]\itemsep0em 
    \item \textbf{Triple classification.} KGs have rich semantics which are represented as triples ($\langle h_u, r, t_v \rangle$). 
    Each triple may be associated with a value to describe its property, such as reliability. 
    By studying head entity $h_u$, tail entity $t_v$ and the relation $r$ between them, we can determine the truth value or the degree of the truth value of unknown triples. 
    \item \textbf{Relation extraction.} The publicly available KGs find use in many real-world applications.
    Despite the success and popularity, these KGs are not exhaustive. 
    Hence, there is a need for approaches that automatically extract knowledge from unstructured text into KGs, namely relation extraction. 
    Specifically, it aims at determining the entailed relation $r$ between a given head entity $h_u$ and tail entity $t_v$ annotated on the text to a background KG.
    \item \textbf{Question answering.} Question answering is a classic computer science research question developed to search for direct and precise answers over the database after understanding users' intentions according to their questions. 
    Current KG-based question answering answers natural language questions with facts from KGs. 
    \item \textbf{Knowledge-aware applications.} Inspired by the richness of the information stored in real-world KGs, researchers propose knowledge-aware models that benefit from integrating heterogeneous information, well-structured ontologies and semantics. 
    Thus, many real-world applications, such as recommendation systems and automatic translation, have been successful in their ability to exploit common sense understanding and reasoning.
\end{itemize}


\section{\revise{Intelligent Drug Discovery}} 
\label{sec:dd}
\begin{table}[!ht]
\caption{Brief explanations of important biomedical terms.
}
\label{table:summary_biomedical_terms}
\setlength{\tabcolsep}{3pt} 
\centering
\footnotesize
\begin{tabular}{|l|l|}
\hline
\textbf{Name} & \textbf{Explanation}  \\
\hline
Disease & 
\specialcell{
Abnormal condition or disorder that affects the normal functioning of the body. 
Diseases can arise from a variety \\ of causes, including genetic mutations, infections, environmental factors, and lifestyle choices.
}
\\
\hline
Drug & \revise{
\specialcell{
A chemical substance designed to interact with the biological systems of the body with the goal of treating, \\ curing, or preventing a disease. 
Drugs can be divided into two categories: \textit{(a)} small molecules, which are organic \\ compounds of small molecular weight and can be of synthetic or natural origin; \textit{(b)} biologics, which can be \\ nucleic acids, proteins, sugars, or living entities, such as antibodies, vaccines, or gene therapy vectors.
}
} 
\\
\hline
Target & \specialcell{
It typically refers to a specific biological molecule, \eg, a protein, enzyme, or receptor, that is involved in the \\ development or progression of a disease and is the focus of a therapeutic intervention. 
Meanwhile, it also can refer \\ to genes or cells in advanced therapy, \eg, gene therapy and cell therapy.
Identifying a target is the starting point \\ for developing new drugs.
} 
\\
\hline
Gene & 
\specialcell{
A gene is a basic unit of heredity encoded by DNA that provides the instructions for making a specific protein. In \\ the context of drug discovery, genes play a crucial role in determining the underlying causes of diseases and \\ identifying potential targets for new therapeutic interventions.
}
\\
\hline
Protein & \specialcell{
A macromolecule involved in numerous functions inside living cells, composed of amino acids.
Proteins can \\ function as enzymes, hormones, structural components, or signalling molecules, and their activity can be disrupted \\ or altered in disease. 
In a pathological context, proteins may be missing, overproduced, or their activity may be \\ altered, leading to the development or progression of a disease.
} 
\\
\hline
Antibody & 
\specialcell{
Protein produced by the immune system in response to foreign substances, such as bacteria and viruses.
\\ Antibodies play a crucial role in the defence against infection and disease, and they are also widely used as drugs \\ in a variety of medical conditions.
}
\\
\hline
Molecule &
\specialcell{
Chemical unit that consists of two or more atoms bonded together. 
The molecular structure of a drug is a critical \\ factor in determining its pharmacological properties and therapeutic potential.
}
\\
\hline
Cell & 
\specialcell{
Cells are the basic unit of life and play a crucial role in many physiological processes. Understanding the behaviour \\ and function of cells is critical in the drug discovery process as it provides important information about the \\ underlying mechanisms of diseases and the potential therapeutic targets for new drugs.
}
\\
\hline
Pathway & \specialcell{
A pathway refers to a series of interconnected biochemical reactions that occur within a cell or organism and lead \\ to a specific physiological or disease state. 
Pathways can be thought of as complex networks of interactions \\ between proteins, enzymes, and other biological molecules, and they play a critical role in many physiological \\ processes, including cellular signalling, metabolism, and gene regulation.
}
\\
\hline
Anatomy & \specialcell{
Study of the structure and organisation of living organisms and the relationship between their parts. 
Anatomy is a \\ critical aspect of drug discovery as it provides important information about the distribution and location of \\ specific tissues, organs, and systems in the body, which can have implications for the efficacy and safety of drugs.
} 
\\
\hline
\end{tabular}%
\end{table}

\begin{regbox} \revise{\textit{\textbf{Fundamentals of Intelligent Drug Discovery}}}
\label{box:fundamentals_drug_design}


\revise{
Different biomedical entities (\Entity) can be classified at different levels, including molecular (\eg, genes (\Gene), molecules (\Molecule)); macro-molecular (\eg, proteins (\Protein), antibodies, enzymes, receptors); cellular (\eg, cells (\Cell)); and organismal levels (\eg, rodents, primates). 
In drug discovery, experiments (called assays) are carried out in vitro (using cells or other microorganisms) and then moved to in vivo (using whole organisms such as rodents, primates, etc.).
}
Table~\ref{table:summary_biomedical_terms} provides brief explanations of important biomedical terms.
Intelligent drug discovery is an interdisciplinary approach that can be embodied in groups of investigations and understanding of the existing entity's \emph{property} and \emph{function} and designing new entities with desired properties or functions. 

\noindent
\textbf{Property} refers to any characteristic or attribute that describes a unique biomedical entity and that can potentially affect its interaction with other entities in biological systems.
Examples of biomedical entity properties include  
\emph{stability} (the ability to maintain its structure and function over time under different conditions), 
\emph{chemical properties} (molecular weight, number and type of functional groups, and chemical reactivity), 
\emph{physical properties} (solubility, partition coefficient, and melting point), 
\emph{pharmacokinetics} (how the substance is absorbed, distributed, metabolised and eliminated in the body), 
\emph{pharmacodynamics} (how the substance affects the body and how the body affects the substance), 
and \emph{toxicity} (how the substance affects the body in the long term and at high doses).

\noindent
\textbf{Function} refers to the specific tasks or activities the biomedical entity performs in the body. 
These tasks or activities can be related to the molecular interactions between the entity and other molecules or the physiological changes caused by the entity in the body.
The function of one entity is its role in the biological system, which is closely relevant to the specific environment and tasks. 
Such as small molecules can function as inhibitors of enzymes that are involved in a particular disease; biologic drugs such as proteins can bind to specific receptors on the surface of cells, thereby triggering a cascade of molecular events that lead to a therapeutic effect; antibodies can be used to target and neutralise specific molecules or cells that are involved in a disease or condition.
\end{regbox}

\begin{figure}[!ht]
\centering
\includegraphics[width=.7\linewidth]{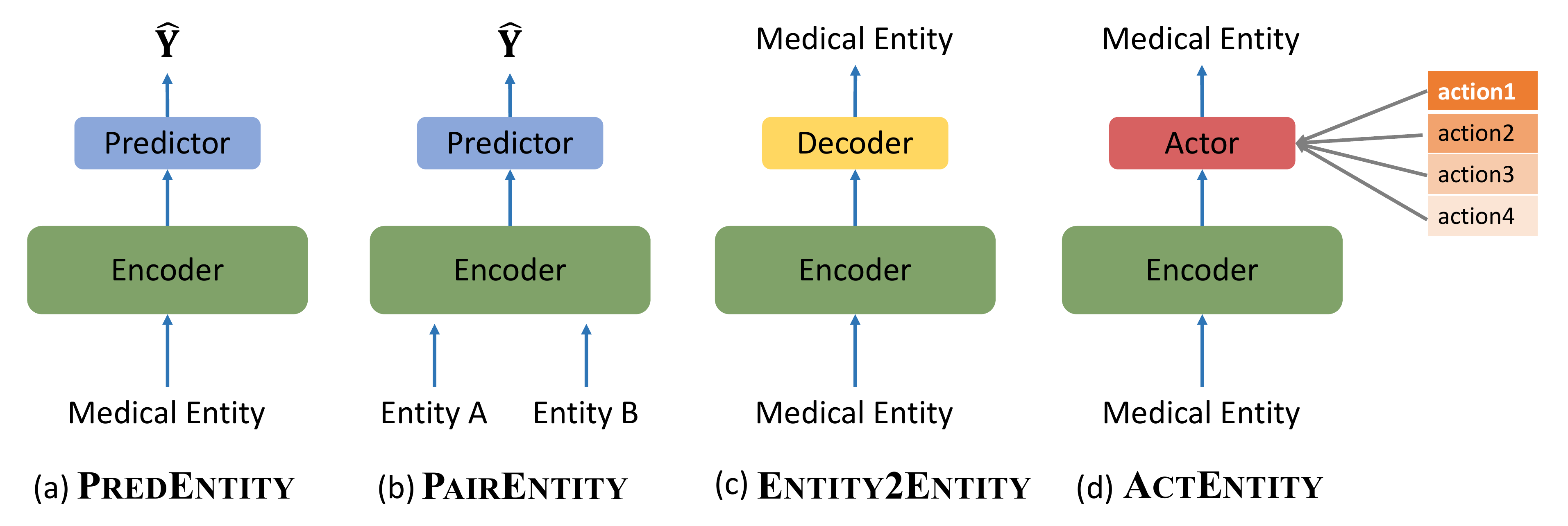}
\caption{
Illustration of the four paradigms in intelligent drug discovery. 
}
\Description{}
\label{fig:artificial_intelligence_drug_design_paradigm}
\vspace{-5mm}
\end{figure}

\subsection{Intelligent Drug Discovery Paradigms}
\label{subsec:intelligent_drug_design_paradigms}


In order to materialise intelligent drug discovery, we describe four paradigms as descriptors for the mainstream applications that we focus on in this paper. 
The illustration of the four paradigms is presented in Figure~\ref{fig:artificial_intelligence_drug_design_paradigm}.
In the following sections, we first describe each drug discovery pipeline and introduce two representative data-driven drug discovery directions, namely GML for drug discovery (Section~\ref{sec:gml_for_dd}) and knowledge-based drug discovery (Section~\ref{sec:k_for_dd}), and corresponding models. 

\begin{itemize}[leftmargin=*]\itemsep0em 
\item 
\textbf{Prediction.}
The Prediction (\Predict\Entity) paradigm consists in predicting a predefined class label or a probability of given biomedical entities. 
It is widely adopted in fundamental drug discovery tasks such as predicting whether a gene sequence generates a valid protein structure~\cite{BBSJORCCK21}, molecular properties~\cite{WAYD21} and missense variant pathogenicity~\cite{ZXFCS22}.
This paradigm can be simply formulated as follows:
\begin{equation}
    \hat{\mathbf{Y}} = \mathrm{Pre}(\mathrm{Enc}(\mathcal{X}))
\end{equation}
where $\hat{\mathbf{Y}}$ can be a value (regression) or a one-hot/multi-hot vector (classification). 
Note that $\mathcal{X}$ can be a an instance of node ($v$), subgraph ($\mathcal{S}$) or graph ($\mathcal{G}$). 
For different types of representations, $\mathrm{Enc}(\cdot)$ can be instantiated as GML models.
$\mathrm{Pre(\cdot)}$ can be implemented as a simple \textit{Multi-layer Perceptron} (MLP)~\cite{R61} or a linear layer, depending on the specific needs. 

\item 
\textbf{Entity pairing.}
The Entity Pairing (\Pair\Entity) paradigm that predicts the interaction between two entities is extensively used in drug discovery tasks, such as molecule-molecule interaction prediction~\cite{HXGZS20,ZLHLZ21}, drug-drug interaction prediction~\cite{MXZW18,DXQXZL20}, protein-protein interaction prediction~\cite{ZPDQSTBLAHMCH12,JDCLB21}. 
This paradigm can be simply formulated as follows:
\begin{equation}
    \hat{\mathbf{Y}} = \mathrm{Pre}(\mathrm{Enc}(\mathcal{X}_a, \mathcal{X}_b))
\end{equation}
where $\mathcal{X}_a$ and $\mathcal{X}_b$ are two nodes ($u, v$), subgraphs ($\mathcal{S}_a, \mathcal{S}_b$) or graphs ($\mathcal{G}_a, \mathcal{G}_b$). 
$\hat{\mathbf{Y}}$ can be discrete (\eg, whether one entity has a pharmacological effect on the other entities) or continuous (\eg, drug similarity). 
$\mathrm{Enc}(\cdot)$ is usually implemented as a GNN~\cite{ZLP23}, while $\mathrm{Pre}(\cdot)$ can be instantiated as an MLP. 

\item 
\textbf{Entity-to-entity.}
\revise{
The Entity-to-Entity (\Entity2\Entity) paradigm can handle a variety of tasks such as molecular generation~\cite{DTDSS18,GWDHSSAHAA18} and protein optimisation~\cite{GMJ18,IGBJ19}. 
}
This paradigm can be formulated as follows:
\begin{equation}
    \mathcal{X}' = \mathrm{Dec}(\mathrm{Enc}(\mathcal{X}))
\end{equation}
where $\mathcal{X}$ is the input entity, such as subgraph ($\mathcal{S}$) or graph ($\mathcal{G}$), and $\mathcal{X}'$ is the predicted entity. 
The encoder module $\mathrm{Enc}(\cdot)$ is usually implemented as a GNN to obtain the corresponding subgraph/graph embeddings, and the decoder module $\mathrm{Dec}(\cdot)$ can be instantiated as an MLP. 

\item 
\textbf{Action prediction.}
The Action Prediction (\Action\Entity) paradigm picks actions to modulate the structure of the target entity that would optimise its properties or functions from an initial state to a terminal state. 
It is widely adopted in fields such as drug generation and optimisation~\cite{YLYPL18,SXGZT20,SXZZZT20}, which are duplicated with the functional field of \Entity2\Entity, but they realise the target with different strategies. 
This paradigm can be formulated as follows:
\begin{equation}
    \mathcal{A} = \mathrm{Cls}(\mathrm{Enc}(\mathcal{X}_{t-1}))
\end{equation}
where $\mathcal{A}=\{a_1, a_2, \dots, a_m \}$ is a sequence of actions, while $\mathcal{X}$ is the input entity such as subgraph ($\mathcal{S}$) or graph ($\mathcal{G}$). 
At each time step $t$, the model predicts an action $a_t$ based on the input entity's state at the previous time step $\mathcal{X}_{t-1}$~\cite{LYJ21}. 
\end{itemize}



\section{Graph Machine Learning for Drug Discovery} 
\label{sec:gml_for_dd}
\begin{figure}[!ht]
\centering
\includegraphics[width=1.\linewidth]{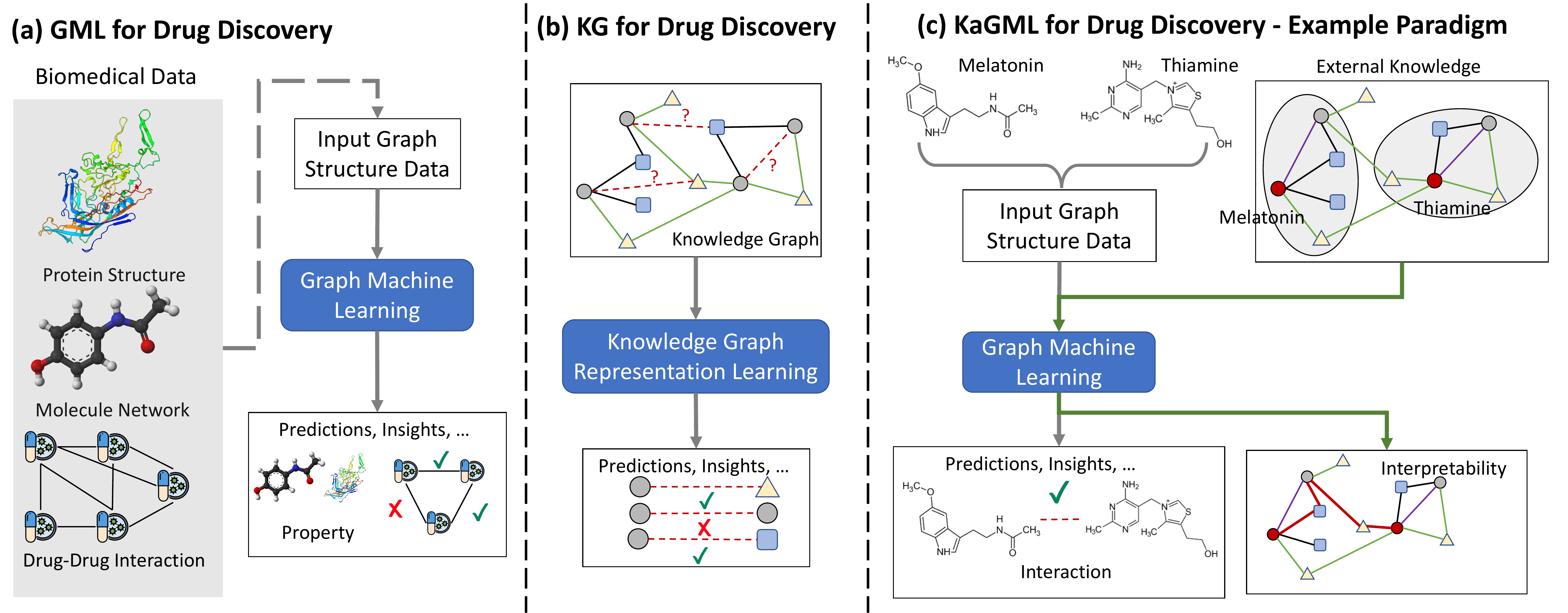}
\caption{
Overview of various intelligent drug discovery approaches. 
(a) GML for drug discovery. 
Biomedical structured data is inputted into a GML model to predict properties or gain insights about the input entities.
(b) KG for drug discovery. 
A KG database is embedded into an embedding space, enabling further analysis to identify any missing relationships between entities.
(c) KaGML for drug discovery. 
Human domain knowledge is integrated into GML models to enhance the flexibility, precision, and interpretability of the drug discovery process.
}
\Description{}
\label{fig:compare_gml_kg_kagml_for_dd}
\end{figure}

Biomedical data are hierarchical, easily organised in ontologies, and interconnected, enabling them to be conveniently represented as graphs and, therefore, highly suitable for consequent use in drug discovery~\cite{GDJSRLHVRTRBBT21,AGS21}.
\revise{
For instance, as shown in Figure~\ref{fig:illustration_of_real_world_biomedical_data}-(a), at the molecular level, atoms can be represented as nodes, and chemical bonds / Euclidean distance as edges of (2D or 3D) molecular graphs~\cite{TLM94,E00}.
}
On the macro-molecule level (protein), interactions (edges) between amino acid residues (nodes) organise as (3D) protein graphs~\cite{Z08,MCSHPZS11}. 
At the compound level, edges in the DDI network can indicate chemical interactions (edges) between drugs (nodes) measured by long-term clinical screens~\cite{PS90,ZZZH09}. 
In these cases, GML, as a powerful graph analysis tool, has been widely applied to a variety of drug discovery tasks (such as those illustrated in Figure~\ref{fig:compare_gml_kg_kagml_for_dd}-(a)). 
Next, we expose representative work following the paradigms defined in Box~\ref{box:fundamentals_drug_design}.

Investigating and understanding the properties or functions of a biomedical entity (\Predict\Entity) is crucial in drug discovery, as it guides researchers to identify potential drug candidates with optimal therapeutic potential, minimal toxicity, and feasibility of synthesis and possibility of patenting. 
Those candidates can be further developed into a safe and effective drug.
Many studies focus on this paradigm, studying the inherent connections between biomedical entities' physical structures and their properties, including molecules, genes, and proteins. 
One mainstream task is interpreting protein functions or molecular phenotypes by modelling protein and molecular graphs~\cite{BOSVSK05,WAYD21,GRKLBVCTFV21,DHB22,SWWRYKZZL22}.
Most of them exploit the 2D structure information, such as graphs constructed based on protein contact map~\cite{GRKLBVCTFV21} or molecular fingerprints and SMILES specification~\cite{WKKGDSL20}\footnote{Some benchmark knowledge, such as protein contact and SMILES specification, are widely applied and we do not consider adopting benchmark knowledge as a factor of being one of KaGML approaches.}. 
Nevertheless, all molecular entities naturally exist in 3D forms in the real world, and their structural conformations determine their function due to the specific intra- and inter-molecular physicochemical interactions~\cite{FWFW20}. 
Molecular 3D structure identification has been a significant open research problem for over 50 years. 
Recently, several computational methods that can successfully predict protein structures have been developed~\cite{LARHZLSVKS22,AEMPDAH22}.
Based on these advancements, a series of works have proposed approaches to model 3D protein and molecular structures to interpret their properties~\cite{BC16,TBSD19,WKKGDSL20,DTME20,C21,JESJTD21,YM22}. 
Compared with previous works that rely on the 2D graph structure, they demonstrate significant superiority in various downstream tasks. 

\revise{
Furthermore, by integrating effective graph structure construction approaches and GML encoder architecture into other paradigms, researchers are able to extend the capacity to more applications. 
Examples of such applications are quantifying entity relations to predict possible interactions (\Pair\Entity)~\cite{FBSB17,LYCJ20,HPMZ22} and modifying or generating new biomedical entities with desired structures expressing proper functions (\Entity2\Entity~\cite{LABG18,IGBJ19,HSVW22,IBCFITWXOB22}, \Action\Entity~\cite{YLYPL18,SXZZZT20,SXGZT20,LJ22,DFSL22}). 
For instance, Fout et al. \cite{FBSB17} represent ligand and receptor proteins as graphs. Using these graph representations generated by GML encoders, they predict the interface between the ligand and receptor proteins. 
Hoogeboom et al. \cite{HSVW22} propose equivariant diffusion GML approaches for 3D molecule generation.
}
For an overview of recent variants and applications of GML for drug discovery, we recommend the comprehensive survey articles~\cite{XHWZL19,WKKGDSL20,GDJSRLHVRTRBBT21,LHZ22}. 


\section{\revise{Knowledge-based Intelligent Drug Discovery}} 
\label{sec:k_for_dd}
Despite the remarkable performance of existing GML models in drug discovery, they are beset by several significant flaws.
These include \textit{a high degree of data dependency}, whereby a significant proportion of performance is contingent on high-quality training data~\cite{M18,PSYTTRSCI19}, as well as \textit{poor generalisation capability}, resulting in uncertain model performance on instances that have never been observed in training data~\cite{BC21}.
To address these limitations, advanced intelligent drug discovery requires robust methods incorporating biomedical knowledge and patient-specific information to generate actionable and trustworthy predictions~\cite{CHZ23,ZTLFS22}.
\revise{
Additionally, a plethora of knowledge has been gained from the successes and failures of the drug discovery and development process. 
With the increasing availability of biomedical data, integrating and extracting relevant information has become a paramount concern.
As a result, numerous knowledge-based intelligent drug discovery approaches have been developed. 
Notably, KGs are fundamental in this domain, as their structures are exceptionally well-suited for integrating heterogeneous knowledge data~\cite{HLHBCHGKB17, ZEPW19, M21, BBYSEBH22}.
}
Section~\ref{subsec:knowledge_representation} systematically introduced the definition and categories of popular knowledge databases and described mainstream applications of the prevalent KGs.
In comparison to the use of GML techniques, the use of human knowledge in drug discovery offers several advantages, such as \textit{(i)} successful integration of a vast amount of knowledge, \textit{(ii)} providing reliable results, and \textit{(iii)} the ability to generate meaningful inferences, which may assist individuals without domain expertise.

The current state-of-the-art approaches in knowledge-based intelligent drug discovery primarily conform to a standard pipeline (as depicted in Figure~\ref{fig:compare_gml_kg_kagml_for_dd}-(b), \Pair\Entity), which entails the acquisition of valid embeddings of nodes and edges in constructed KGs, followed by the prediction of missing edges, corresponding to novel applications~\cite{ZAL18,KCJUBD19,JYCZAL20,HZXOT21,ZTLFS22,BUBJMDCWMK22,GAPSHLF22,GPPPU22}. 
This methodology has been demonstrated in various studies. For instance, \citet{ZAL18} construct a KG of protein-protein interactions, drug-protein target interactions, and polypharmacy side effects, represented as drug-drug interactions, where each side effect is represented as an edge of a distinct type. 
They further propose another noteworthy model, Decagon, to predict the side effects of drug pairs. 
In their seminal work, \citet{JYCZAL20} proposed a groundbreaking heterogeneous information integration method for predicting miRNA-disease associations. 
The methodology consists of constructing a heterogeneous information network by amalgamating known associations among lncRNA, drug, protein, disease, and miRNA. 
Subsequently, the network embedding method GraRep is used to acquire embeddings of nodes in the heterogeneous information network and realise accurate miRNA-disease association prediction. 
For an overview of recent variants and applications of KG for drug discovery, we recommend the comprehensive survey articles~\cite{CHZ23,ZTLFS22}.


\section{Knowledge-augmented Graph Machine Learning for Drug Discovery} 
\label{sec:kagml_for_dd}
%
\revise{Figure~\ref{fig:compare_gml_kg_kagml_for_dd} presents various popular drug discovery mechanisms.}
The techniques relevant to the left figure (\textit{GML for drug discovery}, Box~\ref{box:fundamentals_graph_machine_learning}) and the middle figure (\textit{Knowledge-based intelligent drug discovery}, Box~\ref{box:fundamentals_knowledge_database}) have been briefly discussed, and their important applications in drug discovery have been introduced in Section~\ref{sec:gml_for_dd}-\ref{sec:k_for_dd}. 
\revise{
However, some significant limitations have impeded their application in reality~\cite{GDJSRLHVRTRBBT21,CHZ23}.
On the one hand, the heavy reliance of GML approaches on explicit graph structure and abundant training signals restricts their practical applications as biomedical data generation is time-consuming and costly. 
Additionally, they are aware of restricted biomedical expertise demonstrated in the constructed datasets; nevertheless, advanced knowledge is missing in this context. 
Moreover, existing GML approaches lack interpretability, which is particularly crucial for gaining patients' trust and ameliorating clinical treatment. 
On the other hand, knowledge-based intelligent drug discovery methods demonstrate effectiveness in utilising summarised knowledge. 
Yet, they fail to incorporate the crucial graph structure information of biomedical entities, which significantly restricts their practical effectiveness. 
}
Motivated by these limitations, external biomedical knowledge is extensively incorporated with GML approaches for more efficient drug discovery and development.
We name this novel mechanism \textit{Knowledge-augmented Graph Machine Learning} (KaGML in short), which so far has demonstrated promising results in 
\textit{(i)} achieving more precise drug discovery (Section~\ref{subsec:kagml_preprocessing}-\ref{subsec:kagml_training});
\textit{(ii)} flexibly working with unstructured training data (Section~\ref{subsec:kagml_preprocessing}-\ref{subsec:kagml_pretraining}); 
\textit{(iii)} effectively working with limited training data (Section~\ref{subsec:kagml_pretraining}-\ref{subsec:kagml_training}); 
\textit{(iv)} automatically generating meaningful explanations (Section~\ref{subsec:kagml_interpretability}).

\begin{figure}[!ht]
\centering
\includegraphics[width=.7\linewidth]{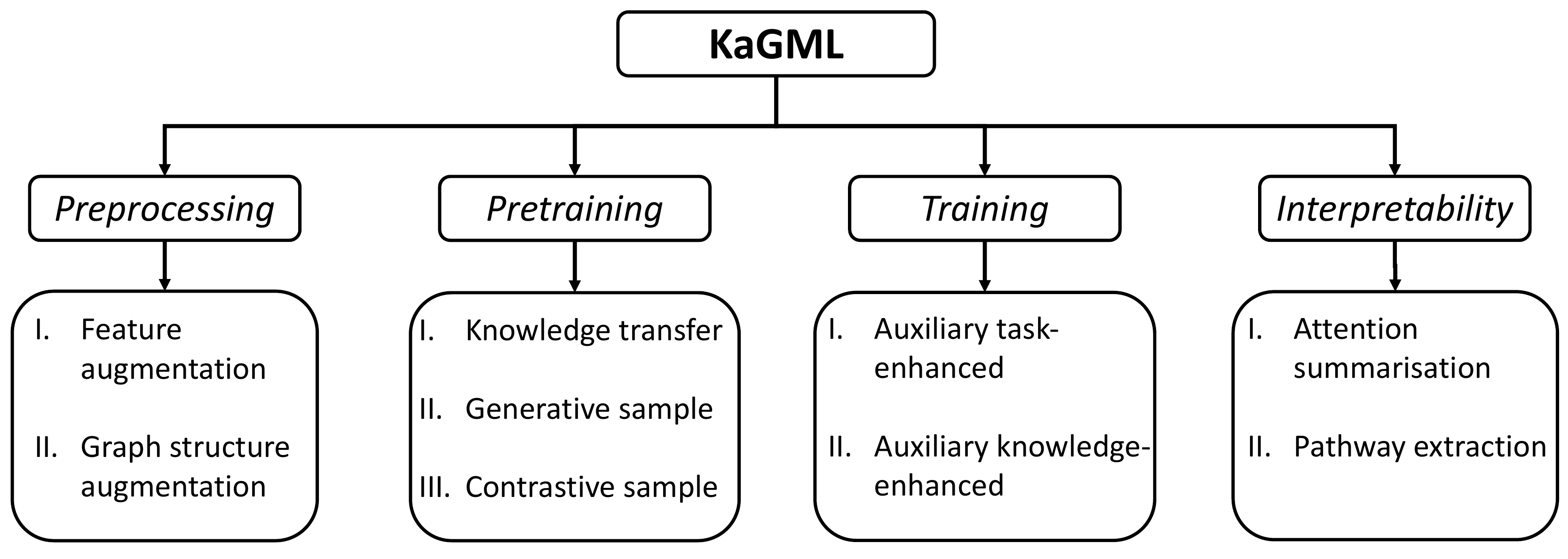}
\caption{
Taxonomy of Knowledge-augmented Graph Machine Learning methods. 
}
\Description{}
\label{fig:kagml_taxonomy}
\end{figure}

The following section provides a comprehensive introduction to the fundamentals of KaGML.
Furthermore, we group KaGML models into four categories, including \textit{preprocessing}, \textit{pretraining}, \textit{training} and \textit{interpretability}, based on the incorporation of external human knowledge into GML models, following the taxonomy illustrated in Figure~\ref{fig:kagml_taxonomy}.
Specifically, \textit{(i)} biomedical knowledge is used to design features or construct graph structure in \textit{preprocessing}, \textit{(ii)} knowledge-augmented \textit{pretraining} strategies are helpful for GML approaches, \textit{(iii)} external knowledge can speed up the \textit{training} process of GML and \textit{(iv)} knowledge databases have been used as a reliable resource to provide meaningful \textit{interpretabilities} of GML models.
The collected work of each category will be discussed in detail.
To facilitate the understanding of the literature, Table~\ref{table:summary_paper} provides a summary of the collected papers that belong to KaGML for drug discovery, which lists their publication information, and the approach used to exploit external knowledge for drug discovery.

\begin{regbox} \textit{\textbf{Fundamentals of Knowledge-augmented Graph Machine Learning for Drug Discovery}}
\label{box:fundamentals_kagml_drug_design}

\noindent
\textbf{Knowledge-augmented Graph Machine Learning.}
The field of Knowledge-augmented Graph Machine Learning (KaGML) concerns the analysis of a graph $\mathcal{G}=(\mathcal{V}, \mathcal{E})$ in combination with a relevant knowledge database $\mathcal{D}$.
The primary objectives of KaGML are twofold. 
The first objective, akin to graph representation learning (Definition~\ref{def:graph_representation_learning}), is to learn a function to generate vector representations for graph elements $f_{KaGML}: (\mathcal{G}, \mathcal{D}) \to \mathbf{Z}$, such that the learned representations can capture the structure and semantics of the graph and the human knowledge stored in the knowledge database. 
The effectiveness of $f_{KaGML}$ is evaluated by applying the vector representations $\mathbf{Z}$ to different downstream tasks (such as defined in Section~\ref{subsubsec:application_graph_machine_learning}).
On the other hand, KaGML models intend to search for direct and precise answers (explanations) over the knowledge database $\mathcal{D}$ to answer users' intentions or explain $f_{KaGML}$'s training and inference processes.
The explanation evaluation process mainly relies on expert manual examination, as shown in the several existing examples in Section~\ref{subsec:kagml_interpretability}. 
We will discuss its limitations and open challenges in Section~\ref{sec:discussion_and_open_challenges}. 
The typical pipeline of KaGML is illustrated in Figure~\ref{fig:compare_gml_kg_kagml_for_dd}-(c). 

The workflow of most machine learning models, including KaGML models, can be divided into four stages: data preprocessing, model pretraining, model training, and model explanation (interpretability). 
Our collected KaGML paper set demonstrates that relevant biomedical knowledge has been successfully applied to all four stages to realise advanced drug discovery.
Next, we will briefly introduce the concept of leveraging KaGML in each stage.

\begin{itemize}[leftmargin=*]\itemsep0em 
    \item 
    \textbf{Preprocessing.}
    This stage includes the construction of the dataset (data integration, transformation, reduction, and cleaning), feature engineering (feature design and selection), etc. 
    \begin{itemize}[leftmargin=*]\itemsep0em
        \item[] In the context of KaGML for drug discovery, this can be summarised as using biological expert knowledge for attribute construction, \ie, designing the node attribute matrix $\mathbf{X}$ and edge attribute matrix $\mathbf{X}^{e}$, and graph structure construction, \ie, constructing the adjacency matrix $\mathbf{A}$.
    \end{itemize}
    
    \item 
    \textbf{Pretraining.}
    Similar to the computer vision and natural language processing communities, the pre-training of GML follows the ``pre-train then fine-tune'' paradigm~\cite{DJVHZTD14,GDDM14,DCLT18}. 
    Specifically, it refers to the setting where a model, initially trained on some tasks, is repurposed for different but related tasks~\cite{HLGZLPL20}. 
    Existing pre-training strategies can be categorised into three categories: multi-task pre-training\cite{WKW16}, generative~\cite{DCLT18} and contrastive~\cite{YCSCWS20} methods according to their design~\cite{LPJZXY21,LZHWMZT23}.
    \begin{itemize}[leftmargin=*]\itemsep0em
        \item[] In the context of drug discovery, finding relevant tasks or labels, and producing generative and contrastive samples are expensive and time-consuming, and incorrect supervision can lead to undermine target tasks. 
        Therefore, KaGML pretraining leverages biological expert knowledge as additional knowledge to pretrain GML models, which can be better applied in downstream tasks after fine-tuning.
    \end{itemize}
    
    \item
    \textbf{Training.}
    This stage refers to the process of learning the trainable parameters $\theta$ (Definition~\ref{def:gml_training}) of $f_{KaGML}$, such that the learned embeddings ($\mathbf{Z}$) can better capture the structure and semantics of graphs and human knowledge databases. 
    \begin{itemize}[leftmargin=*]\itemsep0em
        \item[] In the context of drug discovery, some essential drug discovery information, such as hierarchical descriptions of biomedical entity structure and observed diverse interaction patterns between them, is not included in the target dataset. 
        Therefore, we can alternatively use this information to better train $f_{KaGML}$ by incorporating external knowledge into the mechanism.
    \end{itemize}
    
    \item 
    \textbf{Interpretability.}
    Generating meaningful explanations of machine learning models has been the subject of study in a variety of fields for a long time. 
    \begin{itemize}[leftmargin=*]\itemsep0em
        \item[] In drug discovery, this topic is particularly important since it will significantly affect patients' trust. 
        In addition, timely explanations will help doctors make correct decisions in clinical treatment. 
        KaGML has the ability to generate explanations based on human knowledge and provide better intelligibility. 
    \end{itemize}
\end{itemize}



\end{regbox}

\begin{table}[!ht]
\caption{
Table of collected Knowledge-augmented Graph Machine Learning for drug discovery papers.
Each paper is associated with publication information, support drug discovery task and the process of incorporating external knowledge. 
}
\label{table:summary_paper}
\centering
\setlength{\tabcolsep}{3pt} 
\footnotesize
\begin{tabular}{lllccccc}
\multirow{2}{*}{\textbf{Method}} & \multirow{2}{*}{\textbf{Venue}} & \multirow{2}{*}{\textbf{Year}} & \multirow{2}{*}{\textbf{Task}} & \multicolumn{4}{c}{\textbf{Knowledge Usage Area}} \\
\cmidrule{5-8}
 & & & & \textbf{\textit{Preprocessing}} & \textbf{\textit{Pre-training}} & \textbf{\textit{Training}} & \textbf{\textit{Interpretability}} \\
\midrule
MPNN~\cite{GSRVD17} & ICML & 2017 & \Predict\Molecule & $\checkmark$ & & & \\
D-MPNN~\cite{YSJCEGGHKM19} & J. Chem. Inf. Model. & 2019 & \Predict\Molecule & $\checkmark$ & & & \\
CMPNN~\cite{SZNFLY20} & IJCAI & 2020 & \Predict\Molecule & $\checkmark$ & & & \\
KGNN~\cite{LQWMZ20} & IJCAI & 2020 & \Pair\Drug\Drug & $\checkmark$ & & & \\
MaSIF~\cite{GSMRBBC20} & Nat. Methods & 2020 & \Pair\Protein\Protein & $\checkmark$ & & & \\
KEMPNN~\cite{H21} & ACS Omega & 2021 & \Predict\Molecule & $\checkmark$ & $\checkmark$ & $\checkmark$ &  \\
SumGNN~\cite{YHZGSX21} & Bioinform. & 2021 & \Pair\Drug\Drug & $\checkmark$ & & & $\checkmark$ \\
FraGAT~\cite{ZGZ21} & Bioinform. & 2021 & \Predict\Molecule & $\checkmark$ & & & \\
PAINN~\cite{SUG21} & ICML & 2021 & \Predict\Molecule & $\checkmark$ & & & \\
MDNN~\cite{LGTLZZ21} & IJCAI & 2021 & \Pair\Drug\Drug & $\checkmark$ & & & \\
MoCL~\cite{SXWCZ21} & KDD & 2021 & \Predict\Molecule & & $\checkmark$ & & \\
AlphaFold~\cite{JEPGFRTBZP21} & Nature & 2021 & \Predict\Protein & $\checkmark$ & & $\checkmark$ & \\
KGE\_NFM~\cite{YHYKCCHH21} & Nat. Commun. & 2021 & \Pair\Drug\Target & $\checkmark$ & & & \\
scGCN~\cite{SSZ21} & Nat. Commun. & 2021 & \Predict\Cell & $\checkmark$ & & & \\
MolMapNet~\cite{SZZWQTJC21} & Nat. Mach. Intell. & 2021 & \Predict\Molecule & $\checkmark$ & & & \\
GemNet~\cite{GBG21} & NeurIPS & 2021 & \Predict\Molecule & $\checkmark$ & & & \\
\textsc{HoloProt}~\cite{SBK21} & NeurIPS & 2021 & \Pair\Protein\Protein & $\checkmark$ & & & \\
SynCoor~\cite{GYG21} & NeurIPS & 2021 & \Predict\Molecule & $\checkmark$ & & & \\
KCL~\cite{FZYZDZQCFC22} & AAAI & 2022 & \Predict\Molecule & & $\checkmark$ & & \\
SGNN-EBM~\cite{LQZCT22} & AISTATS & 2022 & \Predict\Molecule & $\checkmark$ & & & \\
scGraph~\cite{YLFZZZJL22} & Bioinform. & 2022 & \Predict\Gene & $\checkmark$ & & & \\
DTI-HETA~\cite{SZWZHB22} & Brief. Bioinform. & 2022 & \Pair\Drug\Target & $\checkmark$ & & & \\
PEMP~\cite{SCMHLMML22} & CIKM & 2022 & \Predict\Molecule & & $\checkmark$ & $\checkmark$ & \\
MISU~\cite{BSR22} & CIKM & 2022 & \Predict\Molecule & & $\checkmark$ & & \\
GraphMVP~\cite{LWLLGT22} & ICLR & 2022 & \Predict\Molecule & & $\checkmark$ & & \\
OntoProtein~\cite{ZBLCHDLZC22} & ICLR & 2022 & \Predict\Protein & & $\checkmark$ & & \\
SphereNet~\cite{LWLLZOJ22} & ICLR & 2022 & \Predict\Molecule & $\checkmark$ & & & \\
\revise{ChIRo~\cite{APC22}} & \revise{ICLR} & \revise{2022} & \revise{\Predict\Molecule} & & & \revise{$\checkmark$} & \\
3DInfoMax~\cite{SBCTDGL22} & ICML & 2022 & \Predict\Molecule & & $\checkmark$ & & \\
DRPreter~\cite{SPBKJ22} & Int. J. Mol. Sci. & 2022 & \Predict\Drug & $\checkmark$ & & & $\checkmark$ \\
DENVIS~\cite{KAPT22} & J. Chem. Inf. Model. & 2022 & \Pair\Protein\Protein & $\checkmark$ & & & \\
ReLMole~\cite{JSLLY22} & J. Chem. Inf. Model. & 2022 & \Pair\Drug\Drug & $\checkmark$ & & & \\
KPGT~\cite{LZZ22} & KDD & 2022 & \Predict\Molecule & & $\checkmark$ & & \\
NequIP~\cite{BMSGMKMSK22} & Nat. Commun. & 2022 & \Predict\Molecule & $\checkmark$ & & & \\
GEM~\cite{FLLHZZWWW22} & Nat. Mach. Intell. & 2022 & \Predict\Molecule & & $\checkmark$ & & \\
ComENet~\cite{WLLLJ22} & NeurIPS & 2022 & \Predict\Molecule & $\checkmark$ & & & \\
\revise{\citet{SDY22}} & \revise{NeurIPS} & \revise{2022} & \revise{\Predict\Molecule} & \revise{$\checkmark$} & & & \\
DTox~\cite{HRM22} & Patterns & 2022 & \Predict\Drug & & & & $\checkmark$ \\
ProteinMPNN~\cite{DABBRMWCDB22} & Science & 2022 & \Action\Protein & $\checkmark$ & & & \\
KEMV~\cite{SZYLZ22} & TKDE & 2022 & \Pair\Drug\Target & $\checkmark$ & & & \\
KG-MTL~\cite{MLSPZ22} & TKDE & 2022 & \Pair\Molecule\Molecule & $\checkmark$ & & $\checkmark$ & \\
\revise{MolKGNN~\cite{LWVMBMD23}} & \revise{AAAI} & \revise{2023} & \revise{\Predict\Molecule} & & & \revise{$\checkmark$} & \\
\revise{ChiENN~\cite{GKTS23}} & \revise{ECML-PKDD} & \revise{2023} & \revise{\Predict\Molecule} & & & \revise{$\checkmark$} & \\
\revise{KeAP~\cite{ZFZCY23}} & \revise{ICLR} & \revise{2023} & \revise{\Pair\Protein\Protein} & & \revise{$\checkmark$} & & \\
\revise{SemiGNN-PPI~\cite{ZQYZGLL23}} & \revise{IJCAI} & \revise{2023} & \revise{\Pair\Protein\Protein} & \revise{$\checkmark$} & & \revise{$\checkmark$} & \\
HIGH-PPI~\cite{GJZJLZYHL23} & Nat. Commun. & 2023 & \Pair\Protein\Protein & $\checkmark$ & & & \\
\revise{MSSL2drug\cite{WCYYLP23}} & \revise{Nat. Mach. Intell.} & \revise{2023} & \revise{\Pair\Drug\Drug} & \revise{$\checkmark$} & & \revise{$\checkmark$} & \\
\revise{KANO\cite{FZZCZSFC23}} & \revise{Nat. Mach. Intell.} & \revise{2023} & \revise{\Pair\Drug\Drug} & \revise{$\checkmark$} & & \revise{$\checkmark$} & \\
\revise{EvolMPNN~\cite{ZM24}} & \revise{ECML-PKDD} & \revise{2024} & \revise{\Predict\Protein} & & & \revise{$\checkmark$} & \\
\revise{GEARS~\cite{RHL24}} & \revise{Nat. Biotechnol} & \revise{2024} & \revise{\Predict\Molecule} & \revise{$\checkmark$} & & \revise{$\checkmark$} & \\
\bottomrule
\end{tabular}
\end{table}

\begin{figure}[!ht]
\centering
\includegraphics[width=1.\linewidth]{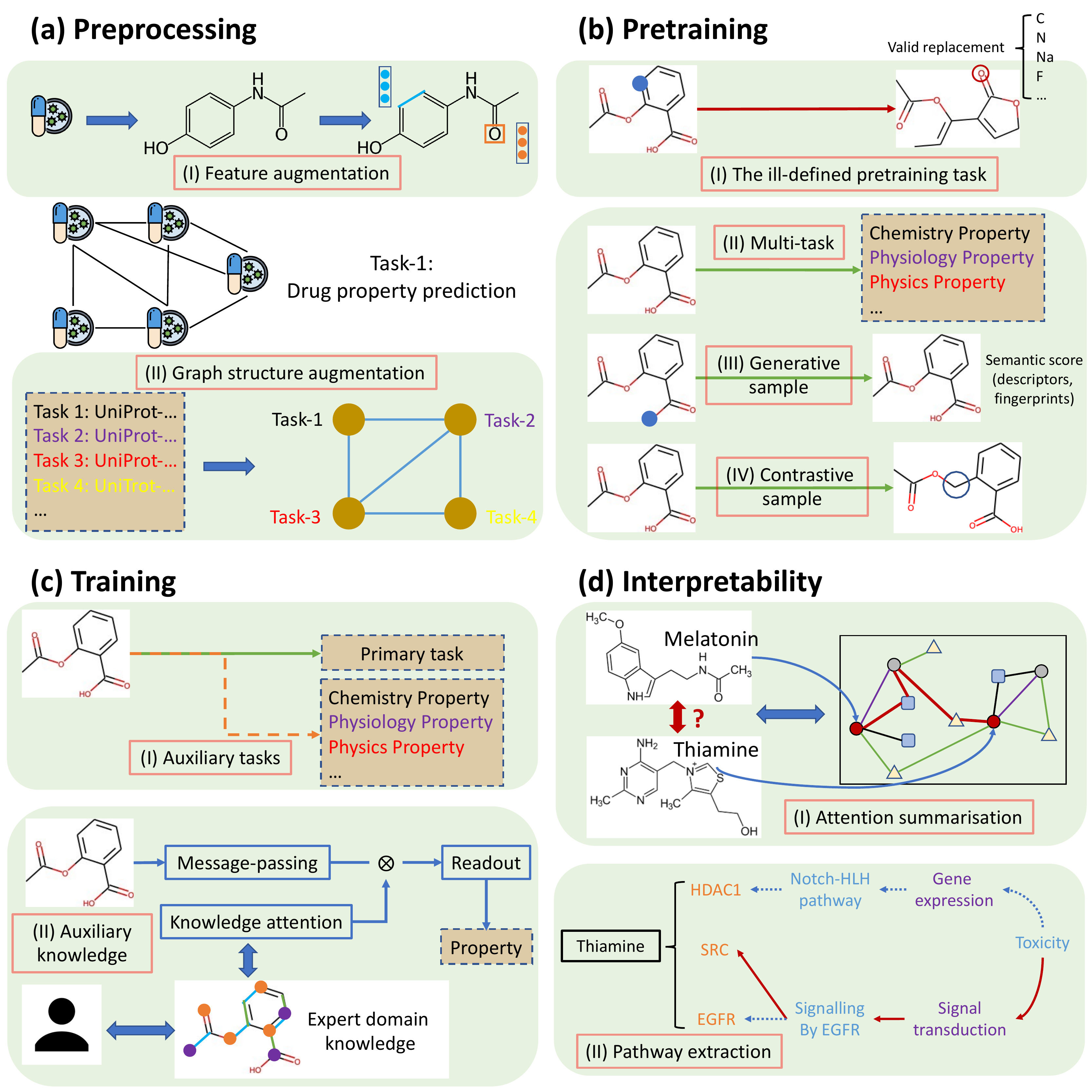}
\caption{
Examples of how knowledge is incorporated in different components of the machine learning pipeline for drug discovery. 
(a) Knowledge is incorporated in the \textit{preprocessing} stage through feature augmentation ((a)-I) and graph structure augmentation ((a)-II).
(b) Knowledge is incorporated in the \textit{pretraining} stage to effectively avoid ill-defined pretraining tasks ((b)-I). 
This can be achieved through techniques such as multi-task guided knowledge transfer pretraining, generative and contrastive sample guided pretraining ((b)-II-IV). 
(c) Knowledge is incorporated in the \textit{training} stage by introducing auxiliary tasks ((c)-I) and auxiliary knowledge ((c)-II)-enhanced training mechanisms. 
(d) Knowledge is incorporated in the \textit{interpretability} stage through techniques such as attention summarisation ((d)-I) and pathway extraction ((d)-II).
}
\Description{}
\label{fig:example_knowledge_usage}
\end{figure}

\subsection{Incorporating Knowledge in \emph{Preprocessing}}
\label{subsec:kagml_preprocessing}
Conventional GML models should ideally be able to extract comprehensive information about a biomedical graph that might be relevant to downstream tasks~\cite{GDJSRLHVRTRBBT21,LHZ22}. However, in practice, there are two limitations that may impede the extraction.
\textit{(i)} GML models require well-defined graph structures (Box~\ref{box:fundamentals_graph_machine_learning}), yet in many real-world scenarios, the graph structure is not explicit and reliable, which precludes the application of GML approaches.
\textit{(ii)} The powerful GML algorithms - GNN models, employ a limited number of message-passing steps, $L$, that is less than the diameter of the biomedical graph, $diam(\mathcal{G})$, meaning that entities that are more than $L$ hops apart will not receive messages about each other.
This results in a representation that is fundamentally localised rather than global, which can hamper the prediction of biomedical properties that are heavily dependent on global features.  
To address these limitations, KaGML approaches incorporate external knowledge into the preprocessing. 

\noindent
\textbf{Feature augmentation.}
The first preprocessing approach, feature augmentation (also known as feature engineering), is a prevalent technique in which external domain knowledge is used to design new features for input data $\mathcal{X}$ in machine learning models. 
This methodology has been extensively studied in the context of intelligent drug discovery, with various studies demonstrating its effectiveness. 
Figure~\ref{fig:example_knowledge_usage}-(a)-I illustrates a typical use case. 
At the molecule level, studies \cite{GSRVD17,YSJCEGGHKM19,SZNFLY20,GYG21,SPBKJ22,BMSGMKMSK22} employ global molecular node and edge features generated by \revise{chemical tools}, such as RDKit~\cite{BHFLGABDL20} and UFF~\cite{RCCGK92}, to enhance vanilla GNN models by enabling them to capture interactions beyond $L$ hops. 
\revise{
These features are a summarisation of theories and informal experimental knowledge. 
}
Additionally, researchers propose various protein backbone features, such as distance between atoms, relative frame orientations and rotations, and backbone dihedral angles, for use in GNN models for 3D molecular graphs~\cite{DABBRMWCDB22}. 
\revise{
Furthermore, the utilisation of both Cartesian~\cite{SUG21,BMSGMKMSK22,FLLHZZWWW22} and Spherical~\cite{GBG21,LWLLZOJ22,WLLLJ22} coordinate systems have been investigated as a scientific laws-based solution to better describe 3D molecular structure.
}
At the macro-molecular level, researchers propose extracting combinations of atomic-level and protein pocket surface-level features for protein pocket modelling~\cite{KAPT22}, and designing sets of strong features to describe the amino-acid type and residue component of amino-acid sequences for protein structure prediction~\cite{JEPGFRTBZP21}. 
\revise{
In contrast to the manual feature design process, \citet{SZZWQTJC21} develop a molecular feature-generation model, MolMap, for mapping molecular representations (MolRs) and fingerprint features (FFs), which is trained by broadly profiling informal experimental knowledge of $8,506,205$ molecules.
}
At the compound level, some studies propose capturing additional domain knowledge from external \revise{biomedical knowledge datasets} (\eg, DrugBank~\cite{WFGLMGSJLS18} and Gene Ontology~\cite{GO21}) to enrich the semantics of entities' embeddings for precise drug-target interaction prediction~\cite{YHZGSX21,LGTLZZ21,SZYLZ22}.

\noindent
\textbf{Graph structure augmentation.}
Graph structure augmentation is another prevalent preprocessing technique in the literature, as illustrated in Figure~\ref{fig:example_knowledge_usage}-(a)-II, in which external domain knowledge is used to reveal relationships between different biomedical entities. 
This approach has been shown to be effective when reliable explicit graph structures are not available. 
For example, various studies have constructed biomedical entity interaction networks based on \revise{external experimental knowledge sources}, such as raw text descriptions~\cite{LQWMZ20}, knowledge databases~\cite{LGTLZZ21,YHYKCCHH21}, ontologies~\cite{SZWZHB22}, pathways~\cite{SPBKJ22}, and entity similarity~\cite{SSZ21}, which reflect implicit connections between target entities.
\revise{
Another noteworthy methodology, as reported by \citet{LQZCT22}, involves constructing a molecular task graph, which is based on the relation of their associated sets of proteins and contains 13\,004 molecules and 382 tasks. 
This method significantly improves the model's generalisation capability. 
}
Additionally, researchers~\cite{ZGZ21,JSLLY22} find that functional molecular graph fragments are useful for capturing meaningful fragment-oriented multi-scale information. 
They construct new fragment-level graphs based on original molecular graphs and learn embeddings that capture semantics at both levels in order to predict target molecule properties. 
Similarly, studies~\cite{GSMRBBC20,SBK21} use the triangulation \revise{software} MSMS~\cite{SOS96} to generate molecular and protein surfaces, which can provide multi-scale structure information for learning representations. 
It is important to note that while incorporating knowledge in preprocessing for intelligent drug discovery has been applied in numerous studies, only representative works of different mechanisms are discussed here, and other relevant works with similar diagrams are omitted. 

\subsection{Incorporating Knowledge in \emph{Pretraining}}
\label{subsec:kagml_pretraining}
One of the main challenges in the use of machine learning techniques for drug discovery is the construction of an appropriate objective function (\ie, defining a learning target) for model training, particularly in the presence of limited supervision labels in biomedical datasets.
One prominent methodology is the pretraining-finetuning pipeline (Box~\ref{box:fundamentals_kagml_drug_design}), where pretrained GNN models have been shown to yield promising performance on downstream drug discovery tasks and have become increasingly popular in the literature~\cite{VFHLBH19,HLGZLPL20,SHVT20,RBXXWHH20,LPJZXY21,LZHWMZT23}.

However, general pretraining strategies may not be well-suited for biomedical graphs.
Unlike word masking in natural languages or data augmentation in images (\eg, resizing and rotation), which do not fundamentally alter the semantics of the raw data, small modifications to biomedical graphs can greatly change the characteristics of the corresponding entities.
For example, as shown in Figure~\ref{fig:example_knowledge_usage}-(b)-I, dropping a carbon atom in the phenyl ring of aspirin can lead to a change in the aromatic system and result in an alkene chain.  
Additionally, without semantics, the relationship between masked nodes and their adjacent nodes is primarily governed by valency rules, which are weaker in comparison to the dependency between neighbouring tokens in natural languages or pixels in images. 
For instance, if an oxygen atom is masked in a molecular graph, the oxygen atom is the expected prediction (as shown in Figure~\ref{fig:example_knowledge_usage}-(b)-I), however, many other atoms can also construct valid molecules according to valency rules.
Thus, enforcing high mutual information between these augmentation pairs may not produce optimal representations for downstream tasks~\cite{SXWCZ21,LZZ22}. 
In other words, previous methods do not adapt to the biomedical scenario. 
To address these challenges, some studies propose knowledge-augmented pretraining strategies. 

\noindent
\textbf{Knowledge transfer pretraining.}
The first methodology is transferring external knowledge to construct pretraining objectives for machine learning models in the field of intelligent drug discovery (Box~\ref{box:fundamentals_kagml_drug_design}). 
One notable example of this approach is the PEMP mode, proposed by \citet{SCMHLMML22}, which identifies a group of useful \revise{(scientific law-, theory- and informal experimental knowledge-based)} chemical and physical properties of molecules, to enhance the capability of GNN models in capturing relevant information through pretraining. 
Furthermore, it is critical to understand the 3D structure of chemical compounds and proteins, as it defines their functions and interactions with other entities. 
GraphMVP~\cite{SBCTDGL22} and 3DInfoMax~\cite{LWLLGT22} adaptively inject the \revise{experimental knowledge} of 3D molecular geometry to a 2D molecular graph encoder such that the downstream tasks on new molecular datasets can benefit from the implicit 3D geometric prior even if there is no 3D information available. 
Additionally, knowledge from large-scale knowledge databases can be exploited as well. 
The OntoProtein framework~\cite{ZBLCHDLZC22} uses knowledge from the Gene Ontology~\cite{GO21} \revise{database} to jointly optimise knowledge and protein embedding during pretraining. 

\noindent
\textbf{Generative-sample pretraining.}
Another widely applied pretraining strategy is relying on generative samples (as outlined in Box~\ref{box:fundamentals_kagml_drug_design}).
This strategy involves training a machine learning model to encode input $\mathcal{X}$ into an explicit vector $\mathbf{Z}$ and subsequently using a decoder to reconstruct $\mathcal{X}$ from $\mathbf{Z}$~\cite{LPJZXY21,LZHWMZT23}. 
Nevertheless, in the context of drug discovery, even small differences in reconstruction can result in significant variations in characteristics. 
As such, it is of significant importance to incorporate external knowledge in order to properly calibrate the pretraining objective. 
For example, \citet{LZZ22} propose the use of molecular \revise{descriptors and fingerprints} as semantic scores and pretrain models by retrieving molecular semantics. 
Similarly, \citet{BSR22} suggest pretraining GNN models by predicting two molecular fingerprints, Morgan fingerprints (also known as extended-connectivity fingerprint ECFP4) \cite{M65,RH10} and MACCS fingerprint~\cite{DLHN02}. 
These fingerprints, which are specifically designed by biomedical experts, are highly relevant to the properties and functions of entities and are sensitive to subtle structural variations. 
The prediction of these objectives enables GNN models to capture crucial factors related to the properties of biomedical entities.

\noindent
\textbf{Contrastive-sample pretraining.}
The third prevalent pretraining strategy is the use of contrastive samples, which involves training a machine learning model to encode input $\mathcal{X}$ into a vector representation, $\mathbf{Z}$, and subsequently using this representation to measure similarity among positive and negative pairs (for example, through techniques such as mutual information maximisation, instance discrimination)~\cite{LPJZXY21,LZHWMZT23}. 
One of the significant challenges in this approach is the generation of adversarial samples, which is particularly relevant in the context of drug discovery~\cite{YCSCWS20}. 
There have been several studies that have proposed effective solutions to this problem.
For example, MoCL~\cite{SXWCZ21} uses domain-specific knowledge, both at the local level (\eg, substructure substitution with bioisostere) and global level (\eg, extended connectivity fingerprints of graphs), to guide the augmentation process and encode the similarity information within the entire dataset. 
Similarly, \citet{FZYZDZQCFC22} construct a chemical element \revise{KG} based on the periodic table of elements (Ptable~\cite{D97}) and enrich original molecular graphs with external chemical knowledge. 
They then propose the KCL model, which designs a knowledge-aware GNN model to encode contrastive molecular graphs that are enhanced with knowledge of atoms and chemical bonds. 
These methodologies successfully incorporate external knowledge by generating contrastive instances.

\subsection{Incorporating Knowledge in \emph{Training}}
\label{subsec:kagml_training}
GML models are a subset of machine learning algorithms that follow the traditional training process and in which a set of parameters are learned through the guidance of supervision signals (Box~\ref{box:fundamentals_graph_machine_learning}).
This process has attracted significant attention in the field of drug discovery, as it plays a crucial role in determining downstream performance. 
However, in this context, the sparsity of training data and the complexity of biomedical process modelling represent significant limitations~\cite{GDJSRLHVRTRBBT21,LHZ22}. 
To address these challenges, recent studies have proposed innovative solutions that incorporate \revise{external knowledge from informal experiments and KGs} to enhance the training process for intelligent drug discovery. 
These solutions are summarised in Figure-\ref{fig:example_knowledge_usage}-(c) and can be broadly grouped into two categories: auxiliary task-enhanced training and auxiliary knowledge-enhanced training. 
The following sections provide detailed descriptions of recent studies in each of these directions.

\noindent
\textbf{Auxiliary task-enhanced training.}
The use of auxiliary tasks in training is a methodology aimed at addressing the issue of limited labelling data in biomedical datasets, as depicted in Figure~\ref{fig:example_knowledge_usage}-(c)-I. 
This approach involves using relevant labelling information of target entities as additional training signals in addition to the primary downstream task labelling data. 
For example, \citet{SCMHLMML22} identify correlations between molecular properties and calculable \revise{theoretical} physics and chemical properties and leverage these physical properties as auxiliary tasks to enhance molecular property predictions.
Similarly, \citet{MLSPZ22} use the embeddings learned from drug-target interaction (DTI) and compound-protein interaction (CPI) as additional knowledge to enhance corresponding molecular embeddings, treating DTI and CPI as auxiliary tasks in training the model for the principal task of molecule-molecule interaction (MMI). 
\citet{JEPGFRTBZP21} adopt multiple sequence alignments (MSA) prediction as an auxiliary task to improve performance in protein structure prediction. 
This methodology, in contrast to knowledge transfer pretraining approaches (\eg, multi-task guided pretraining), requires primary task-related auxiliary labelling data and directly involves them in the model training process.

\noindent
\textbf{Auxiliary knowledge-enhanced training.}
The second knowledge-augmented training methodology is the use of external knowledge to guide the message-passing process, as illustrated in Figure~\ref{fig:example_knowledge_usage}-(c)-II. 
This approach involves leveraging domain knowledge to adjust the internal processes of GNN models.
One example is the \revise{experimental annotation} of atoms, bonds, and functional substructures and their contributions to the properties of molecules, as demonstrated by~\cite{H21}.  
These annotations can be used to guide the attentional message-passing process within the GNN. 
However, the optimal method for adaptively utilising auxiliary knowledge by considering the uncertainty remains an open research question, and we will discuss it in detail in Section~\ref{sec:discussion_and_open_challenges}.

\subsection{Incorporating Knowledge in \emph{Interpretability}}
\label{subsec:kagml_interpretability}
Interpretability, the ability to understand and explain the workings and outputs of a model, has long been a subject of research in various fields and has recently regained popularity due to advancements in the field of machine learning~\cite{TS22}. 
It is not only essential for the development of \textit{eXplainable Artificial Intelligence} (XAI)~\cite{JGS20} but also a crucial step towards \textit{Trustworthy Artificial Intelligence} (TAI)~\cite{TLS21}. 
By providing meaningful and understandable explanations, the model's mechanism and outputs can be better understood by users, thereby increasing trust in the model. 
Furthermore, this issue is of particular importance in the context of drug discovery, where biologists may not have a strong background in computer science. 
However, \textit{a significant limitation of existing intelligent drug discovery approaches is their lack of interpretability}, which hinders the understanding of the model's mechanism and outputs by humans. 
This highlights the need to improve interpretability and trustworthiness from the user perspective.
In response, recent studies have proposed various techniques to explain the model's mechanism and outputs. 
This section specifically focuses on the role of integrating external knowledge for interpretable GML in the context of intelligent drug discovery, as illustrated in Figure~\ref{fig:example_knowledge_usage}-(d).

\noindent
\textbf{Attention summarisation.}
External knowledge in the form of KGs can be used as a means of improving interpretability in GML models for intelligent drug discovery.
One approach, as depicted in Figure~\ref{fig:example_knowledge_usage}-(d)-I, involves the use of attention summarisation methods. 
By utilising machine-readable domain knowledge represented in \revise{KGs}, relevant knowledge can be highlighted for downstream applications.
For example, \citet{YHZGSX21} extract local subgraphs in an external KG in the vicinity of drug pairs to extract biomedical knowledge. 
Through the application of a self-attention mechanism, useful information is captured from the extracted subgraphs to predict drug interactions. 
Additionally, the learned attention weights can be used to prune the subgraphs, thereby generating mechanism pathways for summarising the drug interaction.

\noindent
\textbf{Pathway extraction.}
An alternate category for incorporating knowledge in interpretability, as depicted in Figure~\ref{fig:example_knowledge_usage}-(d)-II, is by using pathway extraction techniques.
Unlike the attention-based knowledge summarisation approach, which captures external knowledge from KGs, this category does not require the construction of \revise{KGs}. 
These techniques adaptively infer pathways pertaining to target biomedical entities from pathway datasets and highlight the key pathways for the explanation.
For instance, \citet{SPBKJ22} introduce cancer-related pathways and construct a cell line network as a set of subgraphs to represent biological mechanisms in detail. 
The extracted self-attention score from the Transformer model can be used to identify which pathways are sensitive to drugs and obtain putative key pathways for the drug mechanism.
Additionally, in another study, \citet{HRM22} design DTox for predicting compound response to toxicity assays. 
They propose a layer-wise relevance propagation approach to infer toxicity pathways of individual compounds from the pathway network.


\section{Practical Resources} 
\label{sec:practical_resources}
\begin{figure}[!ht]
\centering
\includegraphics[width=1.\linewidth]{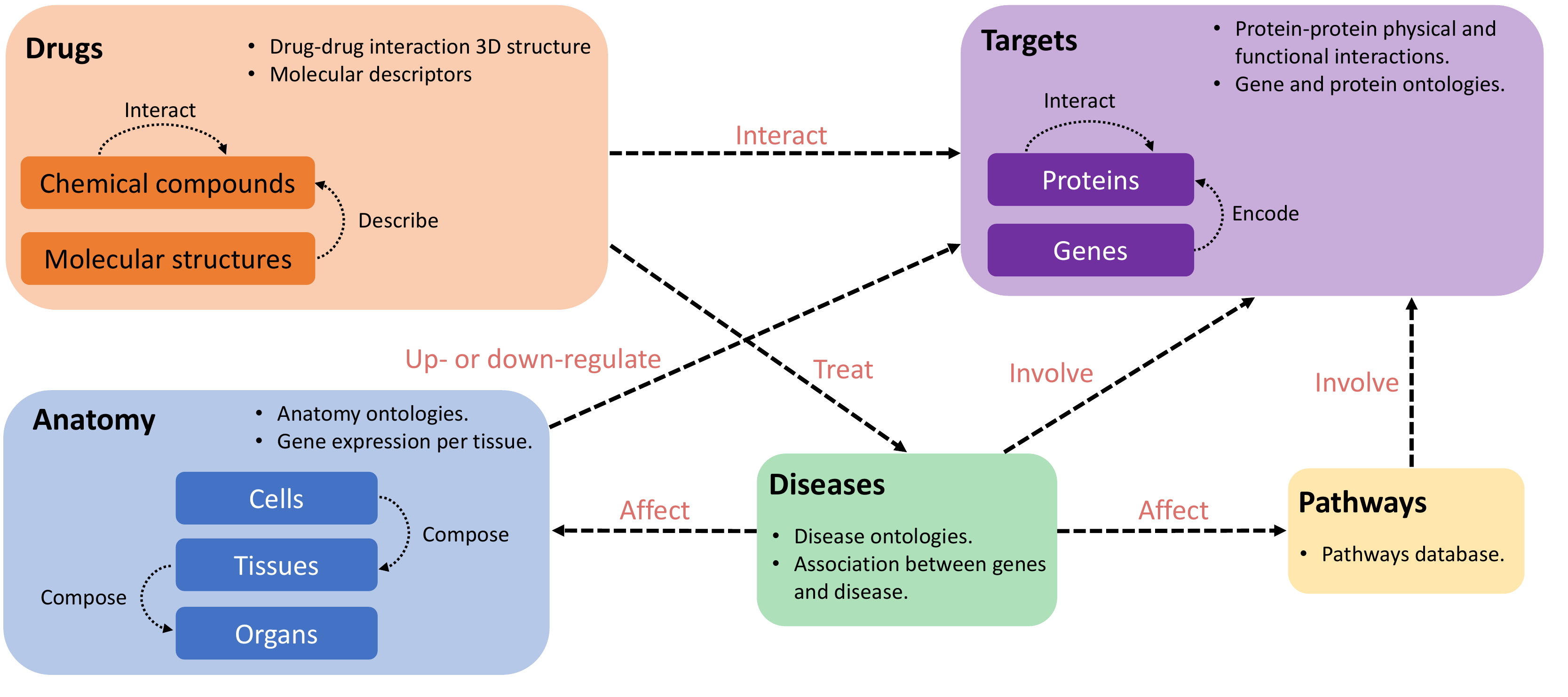}
\caption{
Schematic representation of possible schemata to organise knowledge databases about small molecule drugs from various aspects into one knowledge graph. 
Here we take the example node of small molecule drugs as they are the most used but the ``Drugs'' node could be replaced by another node relative to biological drugs.
}
\Description{}
\label{fig:biomedical_knowledge_databases}
\end{figure}

The current advancements in intelligent drug discovery and development contribute to constructing large-scale knowledge databases describing biomedical entities and patients. 
This enormous amount of information represents an invaluable opportunity to reduce the costs of targeted therapies for intractable diseases while accelerating and explaining the drug discovery process.
Previous sections have systematically introduced background knowledge and state-of-that-art techniques of the novel KaGML paradigm, which can effectively incorporate external biomedical knowledge for intelligent drug discovery. 
A primary factor in the effectiveness of KaGML methods is the availability of qualified knowledge databases that can provide comprehensive and sufficient information.
This section introduces practical resources, including relevant scientific tools and biomedical knowledge databases, that can be used for designing promising KaGML methods for drug discovery and development. 

Depending on the nature of the tasks and the questions asked, different practical resources are required for developing KaGML approaches for drug discovery. 
We also discussed typical research questions that are of great importance for practical drug discovery. 
To facilitate the researchers to conduct relevant studies, we present a large but not exhaustive overview of databases, ontologies and scientific tools that could be useful at any step of drug discovery. 
%

\begin{table}[!ht]
\caption{A collection of \emph{Molecular \& Structural} resources.
}
\label{table:resource_molecular_and_structural_resources}
\setlength{\tabcolsep}{3pt} 
\centering
\footnotesize
\begin{tabular}{|l|l|l|}
\hline
\textbf{Resource} & \textbf{Brief Description} & \textbf{Type} \\
\hline
logP~\cite{WC99} & \specialcell{Measures of a molecule's hydrophobicity, or its partition coefficient between a nonpolar \\ and polar solvent, and is commonly used to predict drug absorption and distribution.} & Formula \\
\hline
rotatable bond~\cite{H21} & \specialcell{Annotation of the (non)rotatable bond.} & Formula \\
\hline 
MolMap~\cite{SZZWQTJC21} & \specialcell{A method to visualise molecular structures in 3D by mapping atomic properties onto a 3D \\ grid, allowing for the exploration and analysis of molecular interactions and properties.} & Software \\
\hline
RDKit~\cite{BHFLGABDL20} & \specialcell{An open-source package to generate chemical features.} & Software \\
\hline
UFF~\cite{RCCGK92} & \specialcell{A molecular mechanics force field designed for the full periodic table.} & Table \\
\hline
Mordred~\cite{MTKT18} & \specialcell{A tool for generating molecular descriptors, which are mathematical representations of \\ molecular structures used for molecular property analysis.} & Software \\
\hline
OpenBabel~\cite{OBJMVH11} & \specialcell{An open-source molecular modelling software that provides a comprehensive toolkit \\ for molecular conversion, visualisation, and analysis.} & Software \\
\hline
MoleculeNet~\cite{WRFGGPLP18} & \specialcell{A benchmark for molecular machine learning, comparing models performances on various \\ molecular property prediction tasks such as solubility, melting point, and binding affinity.} & Database \\
\hline
Ptable~\cite{D97} & \specialcell{A periodic table of chemical elements classified by atomic number, electron configurations, \\ and chemical properties into groups and periods, providing a systematic overview of elements.} & Table \\
\hline
\end{tabular}%
\end{table}

\begin{table}[!ht]
\caption{A collection of \underline{\emph{Compounds}} and \underline{\emph{Drugs \& Targets}} resources.
}
\label{table:resource_compounds_and_drugs_and_targets}
\setlength{\tabcolsep}{3pt} 
\centering
\footnotesize
\begin{tabular}{|l|l|l|}
\hline
\textbf{Resource} & \textbf{Brief Description} & \textbf{Type} \\
\hline
\multicolumn{3}{|c|}{\emph{Compounds}} \\
\hline
CheMBL~\cite{MGBCDFMMMN19} & 
\specialcell{A database of bioactive molecules, assays, and potency information for drug discovery \\ and pharmaceutical research, used to facilitate target identification and selection.}
& Database \\
\hline
PubChem~\cite{KCCGHHLSTY23} & \specialcell{Open database of chemical substances that contains information on their 2D and 3D \\ structures, identifiers, properties, biological activities and occurrence in nature.} & Database \\
\hline
ChEBI~\cite{HODEKMTSMS16} & 
\specialcell{An open-source resource for molecular biology and biochemistry, providing a systematic \\ and standardised vocabulary of molecular entities focused on small chemical compounds.}
& \specialcell{Ontology \\ Database } \\
\hline
\specialcell{KEGG \\ Compound} ~\cite{KFSKI23} & 
\specialcell{A database of small molecular compounds, including their structures, reactions, pathways, \\ and functions, used to provide information on metabolic pathways and cellular processes.}
& Database \\
\hline
DrugBank~\cite{WFGLMGSJLS18} & 
\specialcell{A database includes small molecular compounds, biologics, and natural products, providing \\ information on their properties, mechanisms, and interactions used in drug discovery.}
& Database \\
\hline
\multicolumn{3}{|c|}{\emph{Drugs and Targets}} \\
\hline
DDinter~\cite{XYYWWZWLCL22} & 
\specialcell{A database of protein-protein interactions, providing information on protein targets, their \\ interactions, and related diseases, used to advance drug discovery and development.}
& Database \\
\hline
TCRD~\cite{SMKSNBJVKS21} & \specialcell{Database that aggregates information on proteins targeted by drugs and attributes them a \\ development/druggability level.} & Database \\
\hline
OpenTargets~\cite{OHCSBMLMCF23} & 
\specialcell{A database that integrates diverse genomic and molecular data to provide a comprehensive \\ view of the relationships between diseases, genes, and molecular targets.}
& 
Database
\\
\hline
TTD~\cite{ZZLLWZQC22} & 
\specialcell{A publicly available database that provides information on protein and nucleic acid targets, \\ drugs that target them and related diseases, used to advance drug discovery and development.}
& Database \\
\hline
PharmGKB~\cite{WHGSTWK21} & 
\specialcell{A resource that provides information on the impact of human genetic variation on drug \\ response, used to advance precision and personalised drug therapy.}
& 
Database
\\
\hline
e-TSN~\cite{FSLL22} & \specialcell{A platform that integrates knowledge on disease-target associations used for target \\ identification. These associations were extracted from literature by using NLP techniques.} & \specialcell{Web \\ platform} \\
\hline
nSIDES~\cite{GT22} & 
\specialcell{Multiple resources made available by the Tatonetti lab on drug side effects, drug-drug \\ interactions and pediatric drug safety.} 
& Database \\
\hline
SIDER~\cite{KLJB16} & 
\specialcell{A database of marketed drugs and their side effects, providing information on the frequency, \\ type, and severity of adverse events, used to advance drug safety and pharmacovigilance.}
& Database \\
\hline
\end{tabular}%
\end{table}

\begin{table}[!ht]
\caption{A collection of \underline{\emph{Gene \& Protein}} and \underline{\emph{Pathways}} resources.
}
\label{table:resource_gene_and_protein_and_pathways}
\setlength{\tabcolsep}{3pt} 
\centering
\footnotesize
\begin{tabular}{|l|l|l|}
\hline
\textbf{Resource} & \textbf{Brief Description} & \textbf{Type} \\
\hline
\multicolumn{3}{|c|}{\emph{Genes and Proteins}} \\
\hline
GeneOntology~\cite{GO21} & 
\specialcell{A structured and standardised ontology of gene functions, used to describe and \\ categorise genes and gene products function in a consistent and interoperable manner.}
& Ontology \\
\hline
Entrez~\cite{MOPT05} & 
\specialcell{A database that includes nucleotide and protein sequences, genomic maps, taxonomy, and \\ chemical compounds by referencing other databases, used to query various biomedical data.}
& Database \\
\hline
Ensembl~\cite{CAAAAAAABB22} & 
\specialcell{A database that provides information on annotated genes, multiple sequence alignments \\ and disease for a variety of species, including humans.}
& Database \\
\hline
\specialcell{KEGG \\ Genes}~\cite{KFSKI23} & 
\specialcell{A database that provides information on genes for complete genomes, their associated \\ pathways, and functions in various organisms.}
& Database \\
\hline
BioGRID~\cite{ORCBSWBLKZ21} & \specialcell{A database of protein and genetic interactions curated from high-throughput experimental \\ data sources in a variety of organisms. It includes a tool to create graphs of interactions.} & 
Database
\\
\hline
UniProt~\cite{U23} & \specialcell{A database of protein information, including their sequences, structure, structure and post-\\translational modifications.} & Database \\
\hline
STRING~\cite{SFWFHHSRST15} & 
\specialcell{A database of protein-protein interactions and functional associations, integrating diverse \\ data sources and evidence to provide a weighted network of functional relationships.}
& Database \\
\hline
HumanNet~\cite{HKYKHML19} & 
\specialcell{Network of protein-protein and functional gene interactions, constructed by integrating high\\-throughput datasets and literature, used to advance understanding of disease gene prediction.}
& Database \\
\hline
STITCH~\cite{SSVJBK16} & \specialcell{A database of known and predicted interactions between chemicals and proteins (physical \\ and functional associations), used for the study of molecular interactions.} & Database \\
\hline
PDB~\cite{BWFGBWSB00} & 
\specialcell{A database that provides information on the 3D structure of proteins, nucleic acids, and \\ complex molecular assemblies, obtained experimentally or predicted.}
& Database \\
\hline
RNAcentral~\cite{RNAC21} & 
\specialcell{A repository that integrates information on non-coding RNA sequences for a variety of \\ organisms and attributes them to a unique identifier.}
& Database \\
\hline
\multicolumn{3}{|c|}{\emph{Pathways}} \\
\hline
Reactome~\cite{GJSMRSGSMG22} & 
\specialcell{A database that stores and curates information about the molecular pathways in humans, \\ providing insights into cellular processes and disease mechanisms.}
& Database \\
\hline
\specialcell{KEGG \\ pathways}~\cite{KFSKI23} & 
\specialcell{A database of curated biological pathways and interconnections between them, manually \\represented  as pathway maps of molecular reactions and interactions.}
& Database \\
\hline
WikiPathways~\cite{MARWSHADLE21} & \specialcell{A database of biological pathways that integrates information from several databases, \\ which aims to provide an overview of molecular interactions and reactions.} & Database \\
\hline
\end{tabular}%
\end{table}

\begin{table}[!ht]
\caption{A collection of \emph{Disease} resources.
}
\label{table:resource_disease}
\setlength{\tabcolsep}{3pt} 
\centering
\footnotesize
\begin{tabular}{|l|l|l|}
\hline
\textbf{Resource} & \textbf{Brief Description} & \textbf{Type} \\
\hline
\specialcell{DO~\cite{SMSOMFBJBB22} \\ (Disease Ontology)} & \specialcell{Disease Ontology (DO) is an ontology of human disease that integrates MeSH, ICD, \\ OMIM, NCI Thesaurus and SNOMED nomenclatures.} & Ontology \\
\hline
MonDO~\cite{VMTFHUAABB22} & \specialcell{Semi-automatic unifying terminology between different disease ontologies.} & Ontology \\
\hline
Orphanet~\cite{I97} & \specialcell{A database that maintains information on rare diseases and orphan drugs using cross-references \\ to other commonly used ontologies.} & Database \\
\hline
OMIM~\cite{ABSH19} & \specialcell{A comprehensive, searchable database of gene-disease associations for Mendelian disorders.} & Database \\
\hline
\specialcell{KEGG \\ Disease}~\cite{KFSKI23} & \specialcell{A database of disease entries that are characterised by their perturbants (genetic or \\ environmental factors, drugs, and pathogens).} & Database \\
\hline
ICD-11~\cite{HWJC21} & \specialcell{The 11th version of the international resource for recording health and clinical data in a \\ standardised format that is constantly updated.} & Ontology \\
\hline
Disgenet~\cite{PRSRCSF20} & \specialcell{A database that integrates manually curated data from GWAS studies, animal models, and \\ scientific literature to identify gene-disease associations. It can be used for target \\ identification and prioritisation.} & Database \\
\hline
DISEASES~\cite{PPTBJ15} & \specialcell{A database for disease-gene associations based on manually curated data, cancer mutation \\ data, GWAS, and automatic text mining.} & Database \\
\hline
\specialcell{GWAS \\ Catalog}~\cite{SMABCGGGHH23} & \specialcell{Repository of published Genome-Wide Association Studies (GWAS) for investigating the \\ impact of genomic variants on complex diseases.} & Database \\
\hline
SemMedDB~\cite{KSFRR12} & 
\specialcell{A database that provides information on the relationships between genes and diseases, \\ extracted from the biomedical literature.}
& Database \\
\hline
OncoKB~\cite{CGPKZWRYSN17} & \specialcell{A knowledge precision database containing information on human genetic alterations detected \\in different cancer types.} & Database \\
\hline
HPO~\cite{KGMCLVDBBB21} & \specialcell{The Human Phenotype Ontology (HPO) is an ontology of human phenotypes and database \\ of disease-phenotype associations with cross-references to other relevant databases.} & Ontology \\
\hline
\end{tabular}%
\end{table}

\begin{table}[!ht]
\caption{A collection of \underline{\emph{Medical Terms and Anatomy}} resources.
}
\label{table:resource_medical_terms_and_anatomy}
\setlength{\tabcolsep}{3pt} 
\centering
\footnotesize
\begin{tabular}{|l|l|l|}
\hline
\textbf{Resource} & \textbf{Brief Description} & \textbf{Type} \\
\hline 
Uberon~\cite{HBBBBBCDDD14} & 
\specialcell{A multi-species anatomy ontology. It covers various anatomical systems for organs and tissues.} & Ontology \\
\hline
BRENDA~\cite{CJUHKSNJS21} & \specialcell{A tissue ontology for enzyme source comprising tissues, cell lines, cell types and cultures.} & Ontology \\
\hline
TISSUES~\cite{PSSGJ18} & \specialcell{A database for gene expression in tissues that contains manually curated knowledge, proteo-\\mics, transcriptomics, and automatic text mining. Annotated with BRENDA tissue ontology.} & Database \\
\hline
MeSH~\cite{MJOGPM17} & \specialcell{A comprehensive controlled vocabulary used for biomedical and health-related information.} & \specialcell{Vocabulary} \\
\hline
UMLS~\cite{B04} & 
\specialcell{A biomedical terminologies and ontologies database that integrates and harmonises data \\ from a variety of sources to support clinical documentation and research in healthcare.}
& Ontology \\
\hline
\end{tabular}%
\end{table}

\begin{table}[!ht]
\caption{A collection of publicly available \textit{Knowledge Graphs}.
}
\label{table:resource_publicly_available_knowledge_graphs}
\setlength{\tabcolsep}{3pt} 
\centering
\footnotesize
\begin{tabular}{|l|l|l|}
\hline
\textbf{Resource} & \textbf{Brief Description} & \textbf{Intended Usage} \\
\hline
Hetionet~\cite{HB15} & 
\specialcell{An integrated KG of more than 12,000 nodes representing various \\ biological, medical and social entities and their relationships. It is a \\ valuable resource combining many different databases that can be used \\ for drug discovery and repurposing.}
& \specialcell{Drug discovery \\ Drug repurposing \\ etc.} \\
\hline
PharmKG~\cite{ZRSZXFYN21} & \specialcell{A comprehensive biomedical KG integrating information from various \\ databases, literature, and experiments. It is mainly centered around \\interactions between genes, diseases and drugs.} & Drug discovery \\
\hline
DRKG~\cite{ISMLPZNZK20} & 
\specialcell{A large-scale, cross-domain KG that integrates information about drugs, \\ proteins, diseases, and chemical compounds. It is based on Hetionet, and it \\ was used for drug repurposing for Covid-19.}
& Drug repurposing \\
\hline
CKG~\cite{SCNNSGCAMJ22} & \specialcell{A KG developed for precision medicine that combines various databases \\ and integrates clinical and omics data. It allows for automated upload and \\ integration of new omics data with pre-existing knowledge.} & \specialcell{Biomarker discovery \\ Drug prioritisation.} \\
\hline
OpenBioLink~\cite{BOAS20} & \specialcell{An open-source KG that integrates diverse biomedical data from various \\ databases. It was developed to enable benchmarking of ML algorithms.} & 
\specialcell{Drug discovery}
\\
\hline
BioKG~\cite{WMN22} & 
\specialcell{A KG that integrates information about genes, proteins, diseases, drugs, \\ and other biological entities. It aims at providing a standardised KG in \\  a unified format with stable IDs.}
& \specialcell{Pathway discovery \\ Drug discovery } \\
\hline
Bioteque~\cite{TFBLA22} & 
\specialcell{A KG that enables the discovery of relationships between genes, proteins, \\ diseases, drugs, and other entities, providing an overview of biological \\ knowledge for use in biomedical research and personalised medicine.}
& Broad usage \\
\hline
Harmonizome~\cite{RGFWMMM16} & \specialcell{A KG that focuses on gene- and protein-centric information and their \\ interactions. It provides a unified view of biological knowledge and enables \\ the discovery of new insights fin the biomedical field.} & 
\specialcell{Drug discovery \\ Precision medicine}
\\
\hline
\end{tabular}%
\end{table}

Tables~\ref{table:resource_molecular_and_structural_resources}, \ref{table:resource_compounds_and_drugs_and_targets}, \ref{table:resource_gene_and_protein_and_pathways}, \ref{table:resource_disease} and \ref{table:resource_medical_terms_and_anatomy} summarise different knowledge databases about specific entities. 
Figure~\ref{fig:biomedical_knowledge_databases} describes possible schemes to organise knowledge databases about small molecule drugs from various aspects into one KG, which can be further incorporated into KaGML models for intelligent drug design. 
We chose this example as this type of application is the most widespread, but one could imagine integrating other relevant knowledge in the case of biologic drug development. 
%
Finally, some research groups have combined those resources to give rise to complete KG databases, some of which have a broad line of use, among which are drug discovery and development, that we list in Table~\ref{table:resource_publicly_available_knowledge_graphs}.
Despite these numerous available resources, Section~\ref{sec:discussion_and_open_challenges} will detail the remaining challenges and provide promising future advancements for practical KG construction. 


\section{Discussion and Open Challenges} 
\label{sec:discussion_and_open_challenges}
In this survey, we present a comprehensive overview of the recent advancements in \textit{Knowledge-augmented Graph Machine Learning} (KaGML) for drug discovery. 
To adapt the terminology for researchers from different disciplines, we have systematically introduced essential background knowledge of graph structure data, human knowledge databases and drug discovery.
Afterwards, we present the derived cutting-edge AI techniques, \ie, graph machine learning and knowledge representation learning, and demonstrate their applications for intelligent drug discovery. 

Furthermore, we emphasise the significance of incorporating domain knowledge into the learning process to ensure the precision and interpretability of the models.
In doing so, we introduce the novel mechanism, namely KaGML, and its applications in drug discovery. 
We systematically organise collected KaGML models into four categories with the associated practical resources, including different knowledge databases and knowledge graphs, which are essential to support new KaGML techniques for drug discovery.
KaGML can conveniently incorporate extensive biomedical expertise, which is challenging to include in an individual dataset, into different stages of the GML models. 
These stages include \textit{(i)} designing useful features and constructing explicit graph structure in preprocessing, \textit{(ii)} refining the pretraining process using the properly defined tasks with the guidance of relevant knowledge, \textit{(iii)} ameliorating training process as well as \textit{(iv)} providing understandable explanations towards model inherent mechanism and outputs. 
We are potentially entering a new validation phase for AI techniques within drug discovery and development.
This survey paper can work as a starting point for effective KaGML, with the hope of attracting more research attention.

Despite numerous efforts, the current state of KaGML methods remains limited in terms of practical usability due to certain key hindrances.
In the following paragraphs, the existing deficiencies of KaGML methods will be scrutinised, and potential avenues for future advancements in this field will be highlighted:
\begin{enumerate}[label=(\alph*), leftmargin=*]\itemsep0em 
    \item \textit{Knowledge database composition.}
    A primary factor in the effectiveness of KaGML methods is the availability of qualified knowledge databases that can provide comprehensive and sufficient information.
    While a number of relevant resources have been identified and presented in Section~\ref{sec:practical_resources}, harmonisation and integration of data still pose a significant challenge, as these resources are often diverse, heterogeneous, and distributed across multiple platforms~\cite{SCNNSGCAMJ22}. 
    As such, addressing the lack of standardisation in data integration is a critical area for future research to enhance the power of KaGML~\cite{FSLL22}.
    Such as automatic knowledge integration with diverse types of knowledge, \eg, text, image and graph-structured data. 
    
    \item \textit{Knowledge database compatibility.}
    Many biomedical knowledge databases have to be frequently updated and refined to stay up to date with the current research, which presents a challenge for KaGML methods. 
    To address this, it is recommended that KaGML works to store the versions of the databases used in their experiments for better reproducibility. 
    Additionally, knowledge databases should be managed according to FAIR principles~\cite{WDAAABBBSB16} to enable wider sharing and reuse of existing resources. 
    To further increase their utility for practitioners, it is also recommended that knowledge databases provide clear documentation regarding their resource, schema, and dataset versions. 
    
    \item \textit{Effective knowledge integration.}
    Our survey (Table~\ref{table:summary_paper}) highlights that most existing KaGML methods focus on using biomedical knowledge in the preprocessing stage for augmenting features and graph structure. 
    While this is important, other integration paradigms, such as pretraining, training, and interpretability, hold significant potential for providing improved performance and interoperability, which is particularly critical for precise drug discovery.
    
    \item \textit{Knowledge integration with uncertainty.}
    To date, KaGML works have incorporated external knowledge into preprocessing, pretraining, training, and interpretability for drug discovery. 
    \revise{
    However, these approaches are typically deterministic, ignoring the underlying uncertainty of knowledge and its impact on model learning and inference. 
    For instance, benchmark datasets like MoleculeNet have many flaws and have been criticised in the pharmacology field~\cite{betterddbenchmark2023}. 
    Thus, it is an important area of future research to investigate how to effectively and systematically model knowledge uncertainties for real-world applications.
    }
    
    \item \textit{Advanced interpretability.}
    The enhancement of the interpretability of AI models has the potential to increase the confidence and dependability of patients, as well as to enhance the applicability of the models.
    Section~\ref{subsec:kagml_interpretability} discusses existing KaGML studies, which incorporate external knowledge to achieve fundamental interpretability.
    Nevertheless, there remains a vast scope for further research in the area of advanced interpretability, with the aim of providing more holistic and adaptable explanations, such as advanced reasoning~\cite{ZGZT22,GZRT22} and question-answering capabilities~\cite{CWZZLXG19,WWXHCC19}. 
    
    \item \textit{Careful evaluation benchmark.}
    Designing a comprehensive validation pipeline, such as an explanation verification pipeline~\cite{RBTD19}, is a promising area for future research. 
    While KaGML approaches have been developed to address interpretability problems in drug discovery, the question of how to verify and evaluate the generated explanations remains open. 
    \citet{RZLY22} establish the first five molecular XAI benchmarks to quantitatively assess XAI methods on GNN models and compare their results with those of human experts. 
    Conducting quantitative assessments of XAI methods will accelerate the development of novel approaches and extend their applications.
    
    \item \textit{From drug discovery to more biomedical fields.}
    While this survey focuses on the recent advancements in drug discovery, other fields of biomedical research could benefit from the expanding use of KaGML techniques, including target identification and validation~\cite{MKHGGSAGE19,RB20} and gene and cell therapy~\cite{WMM20,SBNL21}.
    It would be interesting to see the development of a unified KaGML framework that supports diverse healthcare services.
    
    \item \textit{Security and privacy.}
    The advancements in machine learning and growth in computational capacities have transformed the technology landscape but have also raised concerns about security and privacy~\cite{BHZLEB18,PMSW18}.
    Ensuring the security and privacy of KaGML methods for drug discovery is a crucial area of future research, yet it has been largely ignored by existing works. 
    This includes guaranteeing the ownership of knowledge databases, protecting patient-sensitive information, and ensuring the viability of models against malicious attacks.
    For instance, \textit{Federated Learning}~\cite{RHLMRABGLM20,YZXC22} could be an effective solution to collaboratively investigate confidential patient data. 
    
    \item \textit{Efficiency and scalability.}
    Efficiency and scalability are critical factors in large-scale graph analysis~\cite{JPCMP21}.
    The tradeoff between computational efficiency and model expressiveness becomes increasingly challenging as the graph size increases. 
    This challenge is further exacerbated for KaGML. 
    As such, designing efficient and scalable KaGML methods is a critical area for future research that has the potential to enhance the power of KaGML in real-world applications. 
\end{enumerate}


\begin{acks}
This work is supported by the Horizon Europe and Innovation Fund Denmark under the Eureka, Eurostar grant no E115712 - AAVanguard.
\end{acks}

\bibliographystyle{ACM-Reference-Format}
\bibliography{full_format_references}




\end{document}